\documentclass{egpubl}
\usepackage{eg2023}
\usepackage{amsmath}
\usepackage{amsfonts}
\usepackage{multirow}
\usepackage{hhline}
\usepackage{makecell}
\usepackage{wrapfig}
\usepackage{algorithmic}
\usepackage[ruled,vlined]{algorithm2e}
\DeclareMathOperator*{\argmax}{arg\,max}

\usepackage{booktabs} 

\usepackage[ruled]{algorithm2e} 
\usepackage{subcaption}
\SetAlFnt{\small}
\SetAlCapFnt{\small}
\SetAlCapNameFnt{\small}
\SetAlCapHSkip{0pt}
\ConferencePaper        
\usepackage[T1]{fontenc}
\usepackage{dfadobe}  

\usepackage{cite}  
\BibtexOrBiblatex
\electronicVersion
\PrintedOrElectronic
\ifpdf \usepackage[pdftex]{graphicx} \pdfcompresslevel=9
\else \usepackage[dvips]{graphicx} \fi

\usepackage{egweblnk} 

\title[Greedy Shape Curriculum]%
      {Learning to Transfer In-Hand Manipulations Using a Greedy Shape Curriculum}

\author[Yunbo Zhang \& Alexander Clegg  \& Sehoon Ha  \& Greg Turk \& Yuting Ye]
{\parbox{\textwidth}{\centering Yunbo Zhang$^{1}$, Alexander Clegg$^{3}$, Sehoon Ha$^{1}$, Greg Turk$^{1}$,Yuting Ye$^{2}$
    } 
    \\
{\parbox{\textwidth}{\centering $^{1}$Georgia Institute of Technology , $^{2}$Meta Reality Lab, $^{3}$Meta AI Research}
}
}

\begin{document}


\maketitle
\begin{abstract}
In-hand object manipulation is challenging to simulate due to complex contact dynamics, non-repetitive finger gaits, and the need to indirectly control unactuated objects. Further adapting a successful manipulation skill to new objects with different shapes and physical properties is a similarly challenging problem. In this work, we show that natural and robust in-hand manipulation of simple objects in a dynamic simulation can be learned from a high quality motion capture example via deep reinforcement learning with careful designs of the imitation learning problem. We apply our approach on both single-handed and two-handed dexterous manipulations of diverse object shapes and motions. We then demonstrate further adaptation of the example motion to a more complex shape through curriculum learning on intermediate shapes morphed between the source and target object. While a naive curriculum of progressive morphs often falls short, we propose a simple greedy curriculum search algorithm that can successfully apply to a range of objects such as a teapot, bunny, bottle, train, and elephant.



\end{abstract}  
\newcommand{\cmt}[1]{}
\newcommand{\note}[1]{\cmt{Note: #1}}
\newcommand{\todo}[1]{\textcolor{red}{{TODO: #1}}}

\newcommand{\greg}[1]{\textcolor{red}{{Greg: #1}}}
\newcommand{\yunbo}[1]{\textcolor{magenta}{{Yunbo: #1}}}
\newcommand{\sehoon}[1]{\textcolor{orange}{{Sehoon: #1}}}
\newcommand{\alex}[1]{\textcolor{blue}{{Alex: #1}}}
\newcommand{\yuting}[1]{\textcolor{cyan}{{Yuting: #1}}}
\newcommand{\original}[1]{\textcolor{gray}{Original: #1}}
\newcommand{\eqnref}[1]{equation~(\ref{eq:#1})}
\newcommand{\myparagraph}[1]{\noindent\textbf{#1 }}

\newcommand{\revised}[1]{{#1}}

\long\def\ignorethis#1{}

\newcommand{\etal}{{\em{et~al.}\ }}
\newcommand{\eg}{e.g.\ }
\newcommand{\ie}{i.e.\ }
\newcommand{\naive}{{na\"ive }}

\newcommand{\figtodo}[1]{\framebox[0.8\columnwidth]{\rule{0pt}{1in}#1}}
\newcommand{\figref}[1]{Fig.~\ref{fig:#1}}
\newcommand{\secref}[1]{Section~\ref{sec:#1}}
\newcommand{\tabref}[1]{Table~\ref{tab:#1}}
\newcommand{\algref}[1]{Algorithm~\ref{alg:#1}}

\newcommand{\vc}[1]{\ensuremath{\mathbf{#1}}}
\newcommand{\pd}[2]{\ensuremath{\frac{\partial{#1}}{\partial{#2}}}}
\newcommand{\pdd}[3]{\ensuremath{\frac{\partial^2{#1}}{\partial{#2}\,\partial{#3}}}}

\newcommand{\mat}[1]{\ensuremath{\mathbf{#1}}}
\newcommand{\set}[1]{\ensuremath{\mathcal{#1}}}

\newcommand{\vEndEff}{\ensuremath{\vc{d}}}
\newcommand{\vRelMove}{\ensuremath{\vc{r}}}
\newcommand{\sSet}{\ensuremath{S}}

\newcommand{\vControl}{\ensuremath{\vc{u}}}
\newcommand{\vPoint}{\ensuremath{\vc{p}}}
\newcommand{\sSpringCoef}{{\ensuremath{k_{s}}}}
\newcommand{\sDamperCoef}{{\ensuremath{k_{d}}}}
\newcommand{\vHandle}{\ensuremath{\vc{h}}}
\newcommand{\vForce}{\ensuremath{\vc{f}}}

\newcommand{\mTransChain}{\ensuremath{\vc{W}}}
\newcommand{\mRotateTrans}{\ensuremath{\vc{R}}}
\newcommand{\sJoint}{\ensuremath{q}}
\newcommand{\vJoint}{\ensuremath{\vc{q}}}
\newcommand{\mJoint}{\ensuremath{\vc{Q}}}
\newcommand{\mMass}{\ensuremath{\vc{M}}}
\newcommand{\sMass}{\ensuremath{{m}}}
\newcommand{\vGravity}{\ensuremath{\vc{g}}}
\newcommand{\vConstr}{\ensuremath{\vc{C}}}
\newcommand{\sConstr}{\ensuremath{C}}
\newcommand{\vCOM}{\ensuremath{\vc{x}}}
\newcommand{\sGeneralForce}[1]{\ensuremath{Q_{#1}}}
\newcommand{\vStateVar}{\ensuremath{\vc{y}}}
\newcommand{\vControlVar}{\ensuremath{\vc{u}}}
\newcommand{\tr}[1]{\ensuremath{\mathrm{tr}\left(#1\right)}}

%
%

\renewcommand{\choose}[2]{\ensuremath{\left(\begin{array}{c} #1 \\ #2 \end{array} \right )}}

\newcommand{\gauss}[3]{\ensuremath{\mathcal{N}(#1 | #2 ; #3)}}

\newcommand{\pctab}{\hspace{0.2in}}
\newenvironment{pseudocode} {\begin{center} \begin{minipage}{\textwidth}
                             \normalsize \vspace{-2\baselineskip} \begin{tabbing}
                             \pctab \= \pctab \= \pctab \= \pctab \=
                             \pctab \= \pctab \= \pctab \= \pctab \= \\}
                            {\end{tabbing} \vspace{-2\baselineskip}
                             \end{minipage} \end{center}}
\newenvironment{items}      {\begin{list}{$\bullet$}
                              {\setlength{\partopsep}{\parskip}
                                \setlength{\parsep}{\parskip}
                                \setlength{\topsep}{0pt}
                                \setlength{\itemsep}{0pt}
                                \settowidth{\labelwidth}{$\bullet$}
                                \setlength{\labelsep}{1ex}
                                \setlength{\leftmargin}{\labelwidth}
                                \addtolength{\leftmargin}{\labelsep}
                                }
                              }
                            {\end{list}}
\newcommand{\newfun}[3]{\noindent\vspace{0pt}\fbox{\begin{minipage}{3.3truein}\vspace{#1}~ {#3}~\vspace{12pt}\end{minipage}}\vspace{#2}}

\newcommand{\key}{\textbf}
\newcommand{\fun}{\textsc}

\newcommand\high{0.9in}
\newcommand\low{0.6in}



\section{Introduction}


Why do cartoon characters have four digits instead of five \cite{bbc}? It is because the hand is notoriously time-consuming to animate, whether in drawing or digitally, owing to its high degree of articulation. While finger motions can now be tracked by consumer devices as reference \cite{mediapipe,han2020megatrack}, hand-object manipulation remains challenging to capture accurately and conveniently. Moreover, once captured, fixing and editing the result can often be as difficult and tedious as animating from scratch, because both the hand and the object need to move in coordination and convey the physical realism of the interaction. Among all hand-object interactions, in-hand manipulation is the hardest to animate and edit (when compared to other motions such as grasping and pick-n-place), due to frequent contact changes and the dynamic nature of the interaction.

In this work, we apply an imitation learning framework based on deep reinforcement learning (DRL) to physically simulate a reference motion of in-hand object manipulation. We further extend this learning framework to adapt the reference motion to target objects of different shapes and sizes than the source object. As a result, we can capture manipulations of simple primitive objects with the desired styles and actions, and then adapt the hand motion to manipulate more complex objects with various physical properties in a physically realistic way.

\begin{figure}[ht]
\vspace{3mm}
\centering
\includegraphics[width=0.46\textwidth]{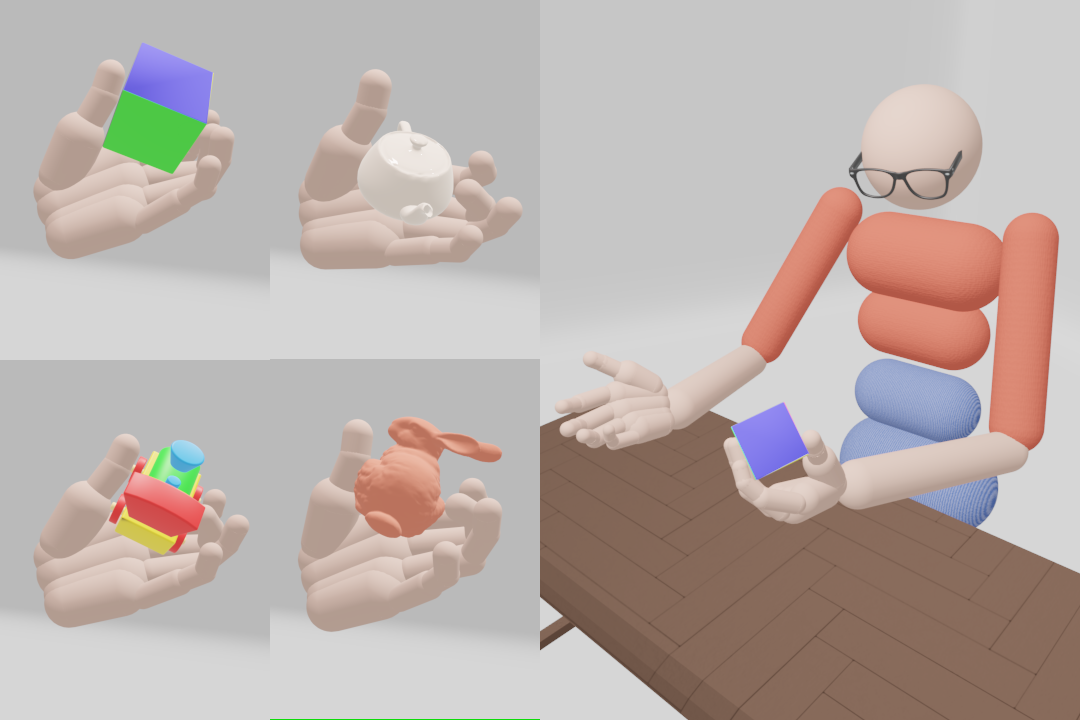}
\caption{Example of motion transfer from a cube to three different objects (left) and a two-hand manipulation (right).}
\label{fig:teaser}
\vspace{-0.5cm}
\end{figure}

Our choice of DRL is inspired by its explosive success in simulating full body motions, including walking, running, jumping, ball bouncing and catching~\cite{won2020scalable,yu2018learning,liu2018learning,merel2020catch}. However, it has not been applied as widely to animations of object manipulation. The robotics community is successful in applying DRL to functional manipulation tasks with manipulators, such as object re-orientation or peg-in-hole \cite{rajeswaran2017learning,akkaya2019solving,chebotar2019closing}, but with less emphasis on naturalness or realism of the hand motion. To achieve natural manipulations, the computer vision community instead captures the kinematic motions of hands interacting with objects from human performance \cite{taheri2020grab,chao2021dexycb,Brahmbhatt_2020_ECCV,brahmbhatt2019contactdb,han2018online}. The resulting datasets enable the synthesis of convincing grasping poses of a rich set of objects \cite{taheri2022goal,wu2021saga,christen2021d,grady2021contactopt}, as well as dexterous and skillful finger maneuvers \cite{zhang2021manipnet}. However, these results are focusing on the kinematic motion and do not enforce proper physical constraints of the interaction. Our work benefits from high-quality and realistic motions of object manipulation, and we further enable physical simulation of the interactions such that our results are responsive to dynamic situations and are physically consistent.

An important challenge of object manipulation stems from the limitless possibility of object shapes. While human hands can effortlessly adapt to geometry variations, control policies have a harder time adjusting. Naive fine-tuning of an existing policy to new shapes is rarely successful. A sensible strategy is then to design a curriculum of increasingly difficult tasks to progressively guide the learning towards solving the target shape. A natural choice is then to use a series of shape morphs between the source and the target objects. Unfortunately, the task difficulty and training progress do not correlate with shape progression, which makes a naive linear curriculum ineffective. Instead, we take inspiration from the work of Wang et al.~\cite{wang2019paired} that generates an open-ended curriculum with increasing complexity by co-evolving the policies and training environments. To this end, we propose a \emph{Greedy Shape Curriculum} across the morph geometry variations. During training, we keep track of the most successful policy for each shape, and pick the most promising shape-policy pair for further training. After a few training epochs, we update the shape-policy pairs by testing the new policy on all shape morphs, then repeat the process. This strategy can effectively adapt an example manipulation of a simple primitive object to a diverse set of everyday objects automatically.

The success of a curriculum hinges on a capable learning framework. Ours is based on the popular  DeepMimic formulation \cite{peng2018deepmimic} on locomotion tasks. When applying it to dexterous manipulations, however, we encountered several unique challenges. For instance, parameters of the contact model can greatly affect both the difficulty and realism of a learned policy. In our case, we carefully chose a set of stiff parameters so the resulting motions do not suffer from sinking or sliding contacts between the hand and the objects, and they work across all our examples without adjustments. We also found it more effective to let the policy output deltas from the reference motion as used in \cite{bergamin2019drecon}, instead of the common choice of direct PD targets \cite{peng2018deepmimic,won2020scalable}. In addition, defining an early termination criterion is more challenging for the control of an unactuated object, because small deviations in the object state could still become unrecoverable. With some deliberate design choices, we arrive at a robust imitation learning framework that can simulate natural and dexterous manipulation of a variety of objects. This framework serves as a solid foundation for the greedy shape curriculum.

Our three main contributions are:
\begin{itemize}
\item We create a framework with well designed simulation and learning configurations for robust dexterous hand manipulation.

\item We propose a novel Greedy Shape Curriculum to automatically transfer an example manipulation of simple shapes to more challenging objects.

\item We showcase that the proposed framework can learn robust policies on various manipulation tasks, even with dynamics variations. 

\end{itemize}




\section{Related Work}

\subsection{Hand Manipulation in Animation}
Generating animation of realistic hand manipulation skills has been a long-studied topic in animation.\revised{Wheatland and colleagues ~\cite{wheatland2015state} summarizes the state of the art in modeling and animation in hand manipulation problems.} Early work exploited traditional optimal control theory to generate hand manipulation skills for rigid objects~\cite{liu2009dextrous,mordatch2012contact,bai2014dexterous} or soft bodies~\cite{bai2016dexterous} by planning contact forces. Instead of solving an optimal control problem, Ye and colleagues~\cite{ye2012synthesis} proposed a randomized sampling algorithm to find a set of diverse strategies.  Zhang~\etal~\cite{zhang2021manipnet} developed a data-driven approach for synthesizing a variety of object manipulation motions via deep learning, but the generated motions are purely kinematic \revised{instead of} simulated with physics. \revised{Reinforcement learning has also been used in generate manipulation animations. Andrews ~\etal ~\cite{andrews2013goal} trains goal directed policies for one-handed manipulation without guidance from motion capture data.} Another line of work has focused on synthesizing grasping motions for simulated hands~\cite{li2007data,christen2021d,hwang2021primitive,pollard2005physically} or whole-body characters~\cite{wu2021saga}. Inspired by prior work, we tackle the problem of hand manipulation by leveraging recent advances in DRL and physics-based simulation.

Our approach of adapting an existing manipulation motion to different objects can also be cast as a motion retargeting problem. Retargeting of complex interactions often considers the spatial relationship of two people \cite{ho2010interaction,jin2018auramesh}, or between the person and the environment \cite{alasqhar2013descriptor}, or through analysis of the object geometry and affordance in the case of manipulations \cite{simeonov2022ndf,taheri2020grab}, without enforcing physical correctness in the results. Recent work starts to utilize geometry representation to improve the learning of manipulation control \cite{she2022highdof}. \revised{More recently, Yang ~\etal ~\cite{yang2022learning} trains policies to control hands to manipulate chopsticks through reconstructing hand motions then then retarget to finish the manipulation tasks.} We instead show that direct imitation learning can be successful in retargeting to different objects with a well designed curriculum. 

\subsection{Robotic Hand Manipulation}
Dexterous hand manipulation has also been an essential topic in robotics. Early work~\cite{huang2000mechanics,sawasaki1991tumbling,han1997dextrous,sundaralingam2018geometric} typically casts manipulation into motion planning problems, which allow robots to achieve various motions, such as tapping, tumbling, or rolling manipulations. However, these motion planning approaches often generate open-loop trajectories which cannot handle dynamic perturbations. This limitation has been overcome by incorporating tactile sensor feedbacks during manipulation~\cite{tahara2010dynamic,li2013rotary}. These methods can allow certain amounts of deviation from the planned trajectory, but they often require accurate dynamic models for both hands and objects. 

Because of hand manipulation's complex and discontinuous nature, researchers have investigated deep reinforcement learning (DRL) algorithms that can solve challenging motor control problems. A common approach~\cite{zhu2019dexterous,nagabandi2020deep,charlesworth2021solving,andrychowicz2020learning,akkaya2019solving} is to cast manipulation into an MDP directly to learn skills without human demonstrations. Andrychowicz~\etal~\cite{andrychowicz2020learning} demonstrated the successful learning of dexterous cube manipulation on a real robot hand by combining pose estimation and deep reinforcement learning. The work is extended to solve a Rubik's cube with a humanoid robot hand by applying proper randomization to system dynamics and designing a curriculum with increasing difficulty. Note that training times for these policies are high, with the Rubik's cube policy requiring several months of training on 64 V100 GPUs and 920 worker machines.

While focusing on a single type of object, other work~\cite{chen2021system,huang2021generalization} trained a control policy in simulation to reorient a wide variation of geometries in hand, including the YCB dataset~\cite{calli2017yale} and various toys. However, this approach of finding a manipulation policy without human demonstrations often leads to unnatural and jerky motion. Many researchers~\cite{gupta2016learning,rajeswaran2017learning,radosavovic2021state,jeong2020learning,kumar2016learning} also have utilized human demonstration to complete dexterous manipulation tasks. For instance, Garcia~\etal~\cite{garcia2020physics} developed a learning framework by estimating the needed contacts to accomplish the tasks on a physics simulator from human motions. Our approach also uses deep reinforcement learning to solve a dexterous hand manipulation problem from the captured hand and object movements.


\subsection{Learning to Imitate Reference Motions}
Since the pioneering work of \revised{Liu~\etal \cite{liu2010sampling,liu2015improving} and }Peng~\etal~\cite{peng2018deepmimic}, learning to imitate a given reference motion has been a promising approach for developing natural and effective physics-based motion controllers. Starting from a short motion clip of walking, running, or kicking, researchers~\cite{peng2018sfv,lee2019scalable,peng2020learning,park2019learning,chentanez2018physics,merel2018neural} have extended the motion imitation framework from various perspectives, including interactivity, data diversity, and character complexity. For instance, researchers trained an interactive controller to complete locomotion tasks from a wide range of walking motions. Won~\etal~\cite{won2020scalable} invented a scalable motion imitation framework by employing an ensemble of expert networks. Lee~\etal~\cite{lee2019scalable} trained a full-body muscular skeleton character to track the motion capture data by introducing a hierarchical controller. Researchers~\cite{won2021control} also developed a technique to solve control of multiple characters in competitive games with a manually designed multi-stage learning. Some prior work~\cite{lee2021learning,won2019learning} have discussed how to adapt policies to variants of the training mocap sequences, rather than simply following the given reference motion. Our work also employs the motion imitation framework while applying it to dexterous hand manipulation, which has \revised{not yet} been fully investigated with motion imitation.

\subsection{Curriculum learning}
Curriculum learning~\cite{bengio2009curriculum} is one of the common techniques for solving challenging problems. It generates tasks with increasing difficulties until it finds solutions for the most difficult scenarios. Many prior works~\cite{karpathy2012curriculum,heess2017emergence,yin2008continuation,tang2021learning} design a curriculum manually using prior knowledge to train locomotion controllers for complex tasks. Instead of a naive curriculum, recent works \cite{won2019learning,xie2020allsteps} propose adaptive curriculum methods based on the value function of the trained policy. In another closely related line of work, researchers \cite{wang2019paired,wang2020enhanced} propose algorithms to design an open-ended curriculum for policies and environments without specifying any target task. In \revised{our} work, we take inspiration from these ideas and propose a new curriculum learning framework, Greedy Shape Curriculum, to generate complex manipulation trajectories with various objects.

\section{Simulation Setup}
We perform physics simulation using Mujoco~\cite{todorov2012mujoco} inspired by successful prior work in hand manipulation~\cite{garcia2020physics,andrychowicz2020learning}. 

\subsection{Motion Capture Sequences}


\begin{wrapfigure}{r}{0.40\columnwidth}
    \centering
    \vspace{-0.25cm}
    \includegraphics[width=0.40\columnwidth]{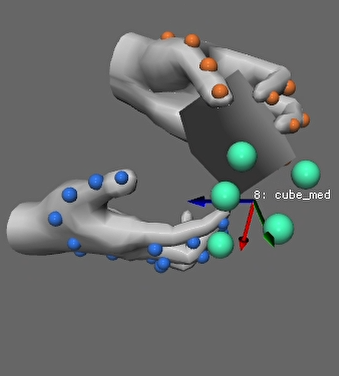}
    \caption{A still frame from an input motion capture sequence.}
    \label{fig:mocap_example} 
    \vspace{-0.15in}
\end{wrapfigure}

We use the object manipulation motion capture dataset from Zhang et al.~\shortcite{zhang2021manipnet}. The data was captured at 120 frames per second with the Optitrack \cite{optitrack} system using reflective markers.  A total of 19 markers are placed on each hand: four on the thumb, three on each finger, and three more on the back of the hand. The Optitrack system outputs object motions as rigid bodies, and the finger motions are reconstructed using DeepLabels \cite{han2018online}.


\subsection{Hand and Object Models}

\begin{wrapfigure}{r}{0.40\columnwidth}
    \centering
    \vspace{-0.25cm}
    \includegraphics[width=0.40\columnwidth]{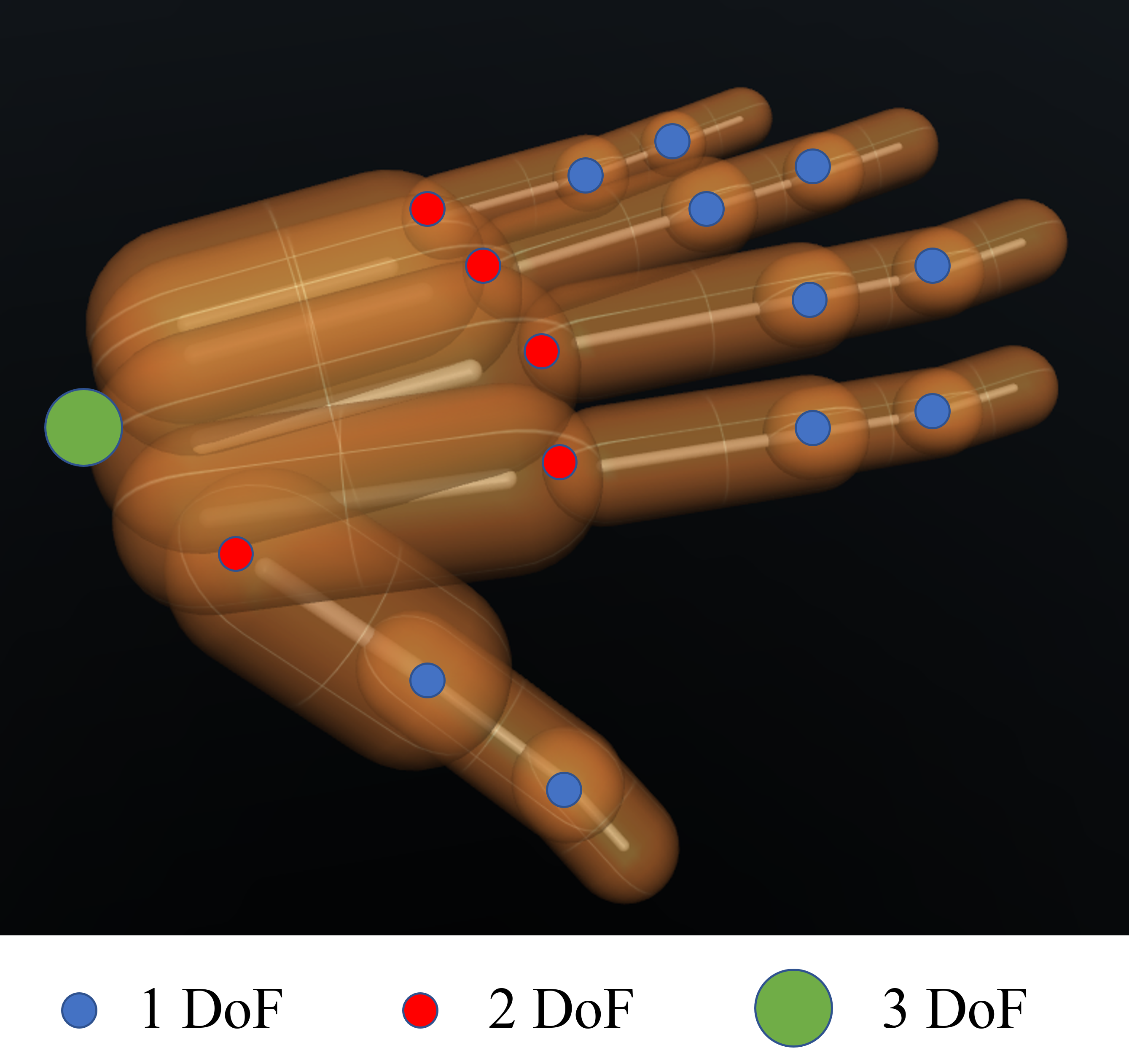}
    \caption{Hand model with $20$ rotational axes.}
    \label{fig:hand_model} 
    \vspace{-0.15in}
\end{wrapfigure}

The dataset comes with a hand model and object models. The hand model consists of $20$ degrees of freedom, and we model each finger segment as capsules (see \revised{Figure}~\ref{fig:hand_model}).  For efficient collision detection in the simulation, objects are modeled either as convex primitive shapes such as a cube or a cylinder, or approximated as a combination of convex primitives if they are nonconvex (e.g. torus and wine glass). \revised{For simplicity purpose, we only enable the collision between the hand and the object. Self collision within the hand are disabled during the manipulation training.}





\subsection{Contact Parameters}
We find that the contact-rich nature of in-hand manipulation tasks demands greater care when choosing simulation parameters related to collision resolution. Mujoco formulates contacts as a constrained optimization problem rather than a linear complementary problem to control the compliance of contact surfaces. In all of our experiments, we choose \revised{Mujoco built-in contact} parameters that correspond to highly stiff surfaces. Specifically, we set the fixed constraint impedance to $0.95$. We also reduce the solver's time constant from $0.02$ to $0.005$, which is the inverse of the natural frequency times the damping ratio. Unfortunately, such stiff contacts increase the instability of the simulator. As a mitigation, we choose elliptic friction cones and the RK4 integrator, and run the simulation at $600$Hz. We additionally found slipping could be a common failure. To increase friction, we set the ratio of friction-to-normal contact impedance to $5$, and strengthen the friction forces with higher normal forces. For more details of the contact model, please refer to the official documentation~\cite{todorov2012mujoco}. We found the above settings empirically works over all the motion sequences we tried, allowing us to achieve effective dexterous manipulation without the fingers ``sinking'' into the object's surface or sliding over sharp edges.

\section{Motion Imitation}
The foundation of our algorithm is an imitation learning framework. In the following sections, we describe the observations, actions, reward functions, and the training procedure. We explain these for a single hand, but our method can trivially incorporate both hands by increasing the state and observation dimensions.

\subsection{Problem Formulation}
Given motion capture trajectories for a hand (or both hands) and an object, we aim to learn an effective policy that can manipulate the object by following a reference trajectory. 

We formulate the control problem as a partially observable Markov Decision Process (PoMDP) with tuple $(\mathcal{S}, \mathcal{O},o, \mathcal{A},\mathcal{T},r, o, \gamma)$, where $\mathcal{S}$ is the state space, $\mathcal{O}$ is the observation space for the hand and the object, $\mathcal{A}$ is the action space for actuating the hand, $\mathcal{T}$ is the transition dynamics (physics simulation), $r$ is the reward function, $o$ is the observation emission function, and $\gamma$ is the discount factor. At a high level, we want to find a parameterized control policy $\pi_{\theta}$ so that it will maximize the expected sum of rewards over a distribution of trajectories
\begin{equation}
    \pi_{\theta^*} = \argmax_{\theta} \mathbb{E}_{(s_0,s_1,\dots,s_T)}\left[\sum_{t=0}^{T}\gamma^{t}r(s_t, \pi_{\theta}(s_t))\right].
\end{equation}

\subsection{Observation representation}
The observation space can be divided into four components: the states of the simulated hand and the object $\{\mathbf{x}_{hand},\mathbf{x}_{obj}\}$, reference states of hand and object from mocap $\{\bar{\mathbf{x}}_{hand},\bar{\mathbf{x}}_{obj}\}$, their differences from simulation $\{\bar{x}_{hand}\ominus x_{hand} ,\bar{x}_{obj}\ominus x_{obj}\}$, and contact information $\{\mathcal{C}\}$. As studied in Bergamin et al.~\shortcite{bergamin2019drecon}, using the difference between simulation and reference as observation improves both training speed and quality. 

\emph{Simulated states:} The positional state of the hand $\mathbf{x}_{hand} =\linebreak[1]  (\mathbf{x}_{h\_root},\linebreak[1]  \mathbf{R}_{h\_root},\linebreak[1]  \cos(q),\linebreak[1]  \sin(q))$ is a $46D$ vector containing the pose of the root and all joint angles of the hand, where $\mathbf{x}_{h\_root} \in \mathbb{R}^3$ is the $3D$ position of the wrist, $\mathbf{R}_{h\_root} \in \mathbb{R}^3$ orientation of the wrist represented as axis-angle, and $q \in \mathbb{R}^{20}$ are all the joint angles in a hand. $\mathbf{x}_{obj} = (\mathbf{x}_{o\_root},\mathbf{R}_{o\_root})$ is a $6D$ vector containing the position and axis-angle orientation of the simulated object. $\mathbf{v}_{hand}$ and $\mathbf{v}_{obj}$ are $26D$ and $6D$ vectors containing the velocities of the hand and the object. Note that we actuate the hand's orientation directly, and we found the use of axis-angle for hand orientation to be especially important for success.

\emph{Reference states:} $\bar{\mathbf{x}}_{hand}$ and $\bar{\mathbf{x}}_{obj}$ are the reference poses of the hand and object at the current frame expressed in the same format as the pose of simulated hand and object. 

\emph{State differences:} $\bar{x}_{hand}\ominus x_{hand}$ and $\bar{x}_{obj}\ominus x_{obj}$ are the differences between the simulated pose and the reference pose of both the hand and object. For all the rotational information, we evaluate the rotation differences in $SO(3)$ and express the difference into the corresponding format \revised{of axis angle or sine and cosine of Euler angles} used in the observation.

\emph{Contact information:} We include an additional $19D$ vector $\mathcal{C}$ to capture the contact forces between the hand and the object. We surround each finger capsule with a contact sensor that is a slightly larger capsule with a 10\% larger radius. These contact sensors register contact forces from the object at each control step, and we record the sum of all contact forces exerted by the object on each rigid segments of the hand through the contact sensors. Because the sensors correspond to finger segments, their readings indicate how much contact forces are being exerted, and implicitly inform where the contacts are. 

Combining all the described observation components, we have a $207D$ vector that describes the state of the simulation when we are considering a manipulating task involving a single hand. When we train the policy to track a two-hand manipulation sequence, we double the observations for the hand and end up with a $390D$ vector.

\subsection{Action Representation}
\begin{figure}
\centering
\includegraphics[width=0.48\textwidth]{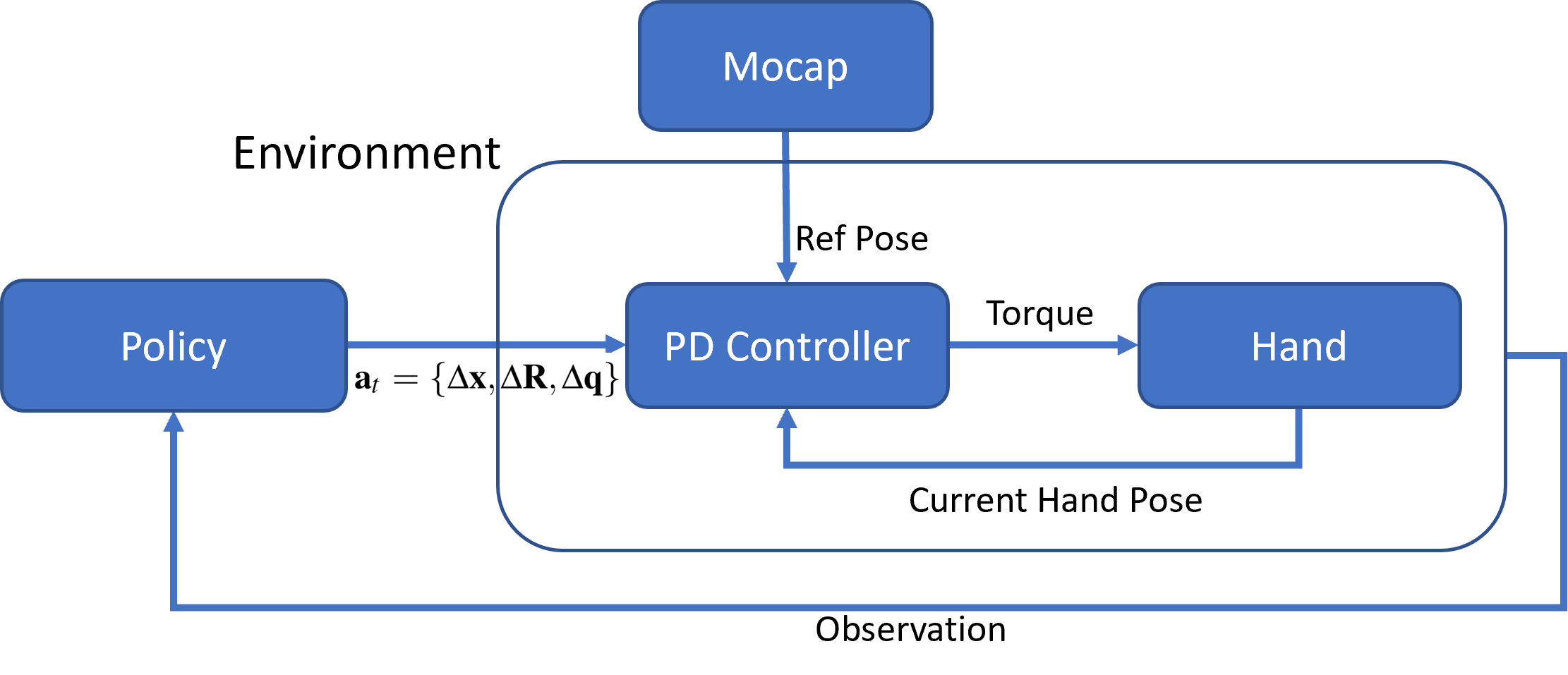}
\caption{Overview of the control loop}
\label{fig:control_loop}
\end{figure}

Similar to the approach in \cite{bergamin2019drecon}, at each time step $t$, the policy outputs an action $\mathbf{a}_t = \{\Delta{\mathbf{x}},\Delta{\mathbf{R}},\Delta{\mathbf{q}}\}$, a $26D$ vector specifying the spatial displacement to the hand's reference pose, where $\Delta{\mathbf{x}} \in \mathbb{R}^3$ is the hand root linear displacement, $\Delta{\mathbf{R}} \in \mathbb{R}^3$ is the root orientation displacement expressed in axis-angle, and $\Delta{\mathbf{q}} \in \mathbb{R}^{20}$ is the joint angle displacement for each joint. We apply an exponential action filter with $\alpha=0.3$ to generate smoother motions. Once the PD target is computed, we compute joint torques using the stable-PD controller~\cite{tan2011stable}. \revised{The full control loop is shown in Figure \ref{fig:control_loop}}



\subsection{Reward Function}
Our goal is to track the reference motions of both the hands and the object as closely as possible. Inspired by the original DeepMimic paper~\cite{peng2018deepmimic}, we design our reward function as follows:
\begin{equation}
    r = w_{od}r_{od}+w_{or}r_{or}+w_{hd}r_{hd}+w_{hr}r_{hr}+w_{hj}r_{hj}
\end{equation}
which consists of the object position term $r_{od}$, the object rotation term $r_{or}$, the hand position term $r_{hd}$, the hand orientation term $r_{hr}$, and the hand joint term $r_{hj}$.
To enforce a match between the simulated object and the reference object's position and orientation, we define the terms $r_{od}$ and $r_{or}$:
\begin{equation}
    r_{od} = \exp\left(-k_{od}\|\hat{x}_{obj}-x_{obj}\|^2\right),
\end{equation} and
\begin{equation}
    r_{or}=\exp\left(-k_{or}\|\hat{q}_{obj}^{-1}q_{obj}\|^2\right),
\end{equation}
which compares the object's position $x_{obj}$ and orientation $q_{obj}$ to their desired values.
For all $N$ rigid segment of the hand with the index i, we define the reward terms $r_{hd}$ and $r_{hr}$:
\begin{equation}
    r_{hd}=\exp\left(-k_{hd}\sum_{i =1}^N\|\hat{x}_{i}-x_{i}\|^2\right),
\end{equation} and 
\begin{equation}
    r_{hr}=\exp\left(-k_{hr}\sum_{i =1}^N\|\hat{q}_{i}^{-1}q_{i}\|^2\right),
\end{equation}
where $x_{i}$ and $q_{i}$ represent the position and orientation of the $i$th body segment.
In addition enforcing the hand rigid segment's tracking, we also define the reward term $r_{hj}$
\begin{equation}
    r_{hj}=\exp\left(-k_{hj}\sum_{i =1}^{M}\|\hat{\theta}_{i}-\theta_{i}\|^2\right),
\end{equation}
by comparing all the current and desired joint angles, $\theta$ and $\hat{\theta}$. For all experiments, we set the weights as $w_{od}=4$, $w_{or}=4$, $w_{hd}=0.05$, $w_{hr}=0.05$, and $w_{hj}=0.1$. 

\subsection{Terminal Condition}
As studied in DeepMimic~\cite{peng2018deepmimic}, early termination of a rollout when the simulation enters an unrecoverable state can save computation on low value trajectories. We design the early termination criteria to restrict how much the object's state is allowed to deviate from the reference: either $d_{thr}=10cm$ in translation or $\phi_{thr}=60^\circ$ in rotation. We choose these thresholds to allow the hands to explore its action space more freely, but this still eliminates irredeemable failures.  



\section{Greedy Shape Curriculum for Novel Objects}

It would be undesirable to record a new motion capture sequence for each new object that we want to manipulate. Instead, we would like to generalize an existing motion example to different objects in simulation. For example, we may want to manipulate a teapot or a toy train using the same reference motion for a cube. However, it is not straightforward to learn an effective policy for a new shape because it often requires significant changes in the control strategy.

Our key intuition is that we can co-train policies on a set of intermediate shapes morphing between the original object and the target object as a curriculum. A naive method would be to use a training curriculum that starts the learning from the source object and gradually morph the shape to the target in a linear progression. In practice, however, this linear curriculum is often unsuccessful because the morphing progression may not exactly correlate to the task's difficulty. \revised{A better tuned morphing algorithm might be able to give stronger correlation between morphing progression and training difficulty, but such algorithm requires additional human effort, and may not be able to generalize across different source target pairs.}

\begin{figure}
\centering
\includegraphics[width=0.48\textwidth]{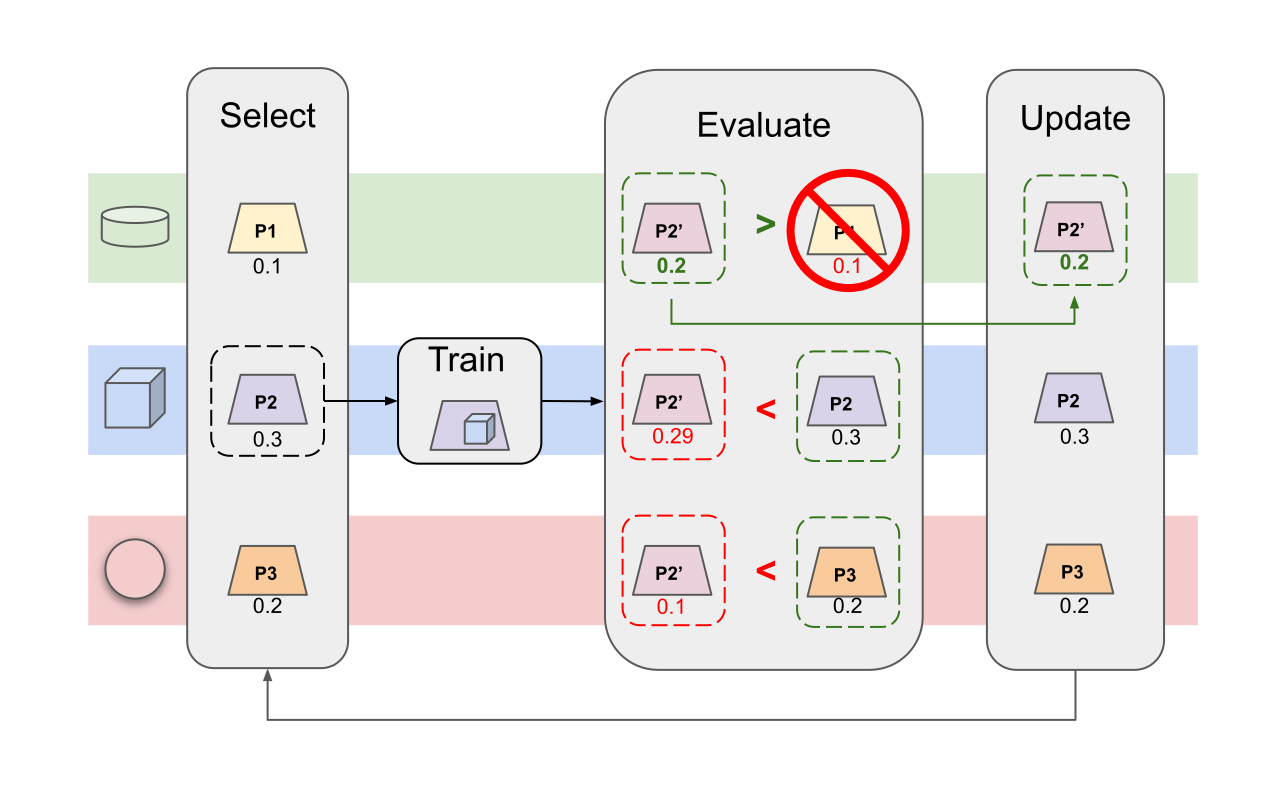}
\caption{Illustration of our greedy shape curriculum. Each iteration of the algorithm (1) selects and trains the most promising (policy, shape) pair, (2) evaluates the updated policy on all shapes, and (3) overwrites a shape’s policy pairing if the new policy is better than the cached policy. Example policy performance metrics are displayed numerically below each policy shape.}
\label{fig:morph_training}
\end{figure}
Instead, we design a novel training schedule that allows greedily switching between any intermediate shape morphs for a more flexible curriculum. Our algorithm maintains a collection of best policies for each shape. For every $K=20$ policy iteration, it selects the best performing \emph{unsuccessful} morph and its paired policy for the next round of training. Once the policy is further trained, the newly updated policy's performance is evaluated across the entire collection of shapes, and overrides existing policies if it performs better (Figure \ref{fig:morph_training}). Despite its greedy nature, we found this automated curriculum more effective in policy transfer than naive fining-tuning or a fix curriculum. The full procedure is described in Algorithm~\ref{alg:shape_morph_training}.
\begin{figure*}[h!]
    \centering
    \includegraphics[height=\low]{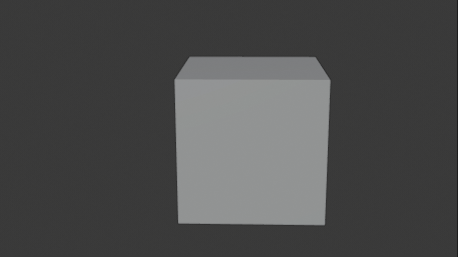}
    \hfill
    \includegraphics[height=\low]{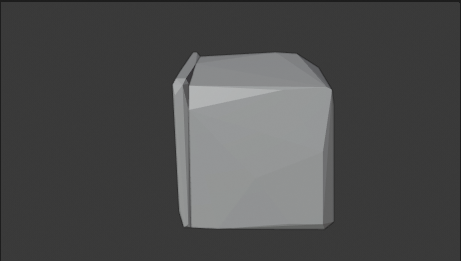}
    \hfill
    \includegraphics[height=\low]{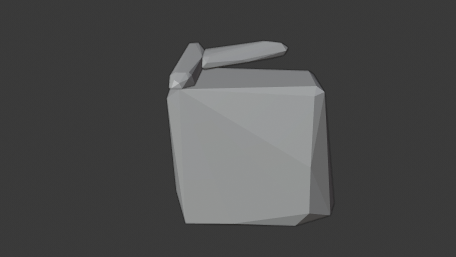}
    \hfill
    \includegraphics[height=\low]{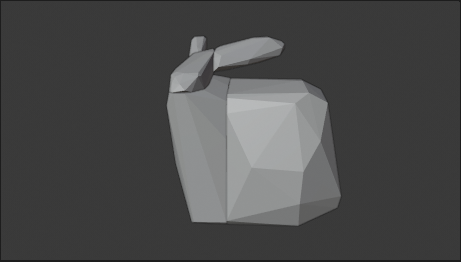}
    \hfill
    \includegraphics[height=\low]{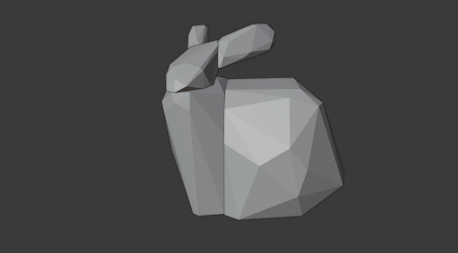}
    \hfill
    \includegraphics[height=\low]{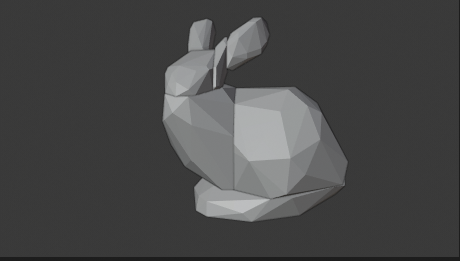}
    \hfill
    \caption{Morph stages of the collision shapes for transferring the cube motion to a bunny after applying V-HACD.}
    \label{fig:collision_shape_morphs}
\end{figure*} 
\begin{figure*}[h!]
    \centering
    \includegraphics[height=\high]{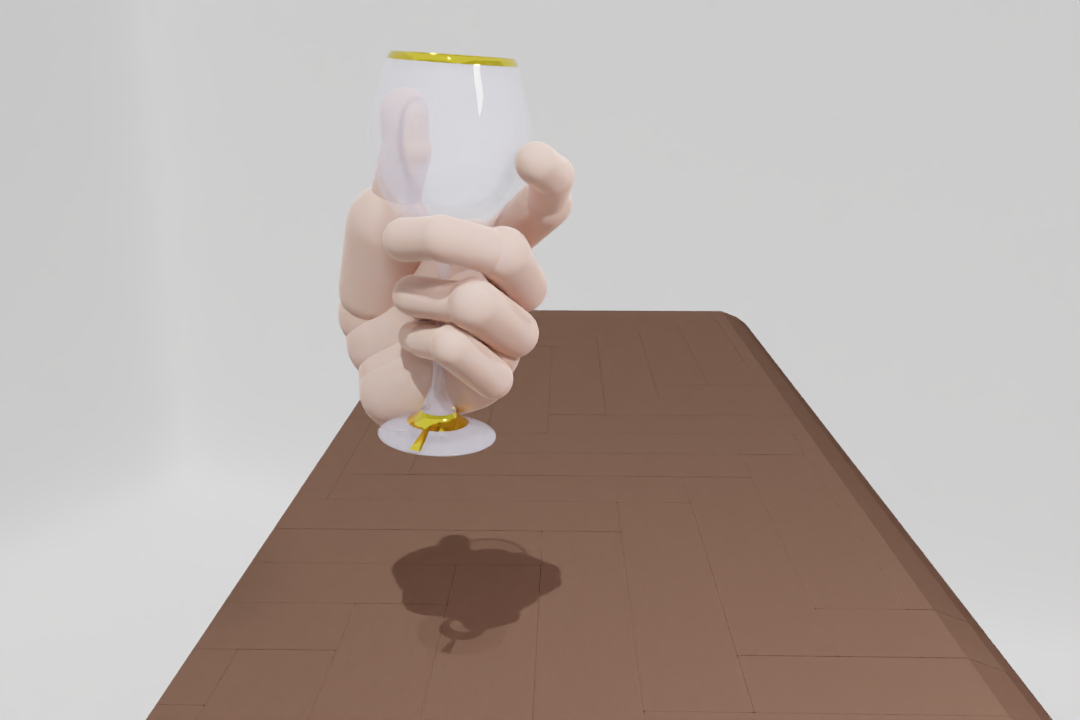}
    \hfill
    \includegraphics[height=\high]{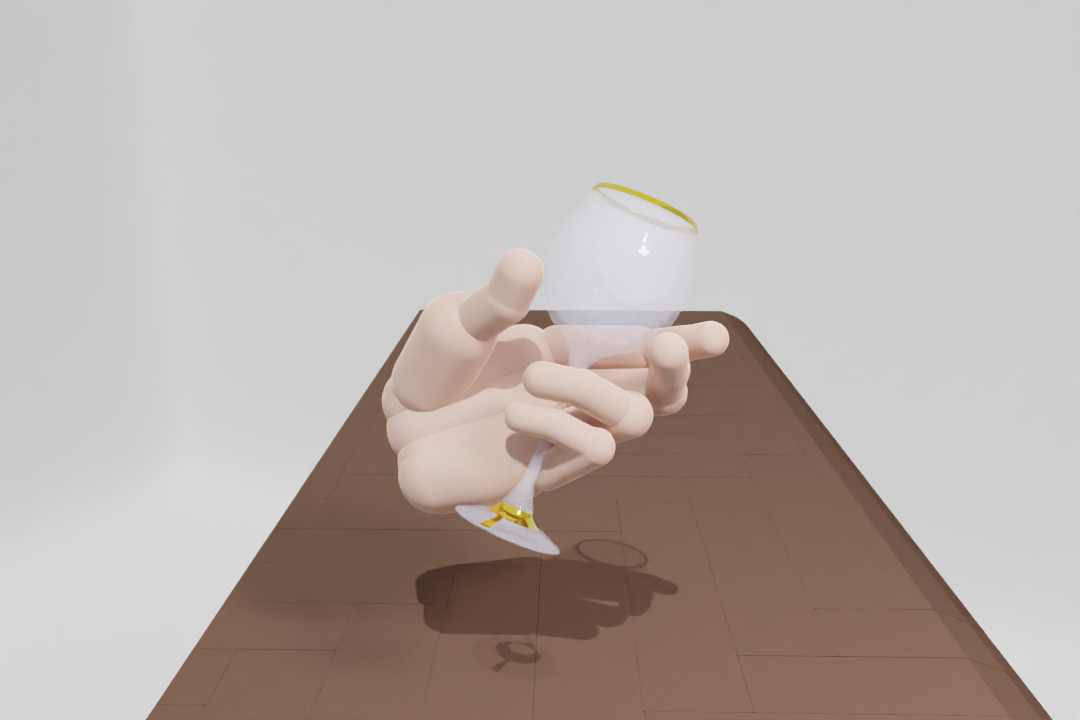}
    \hfill
    \includegraphics[height=\high]{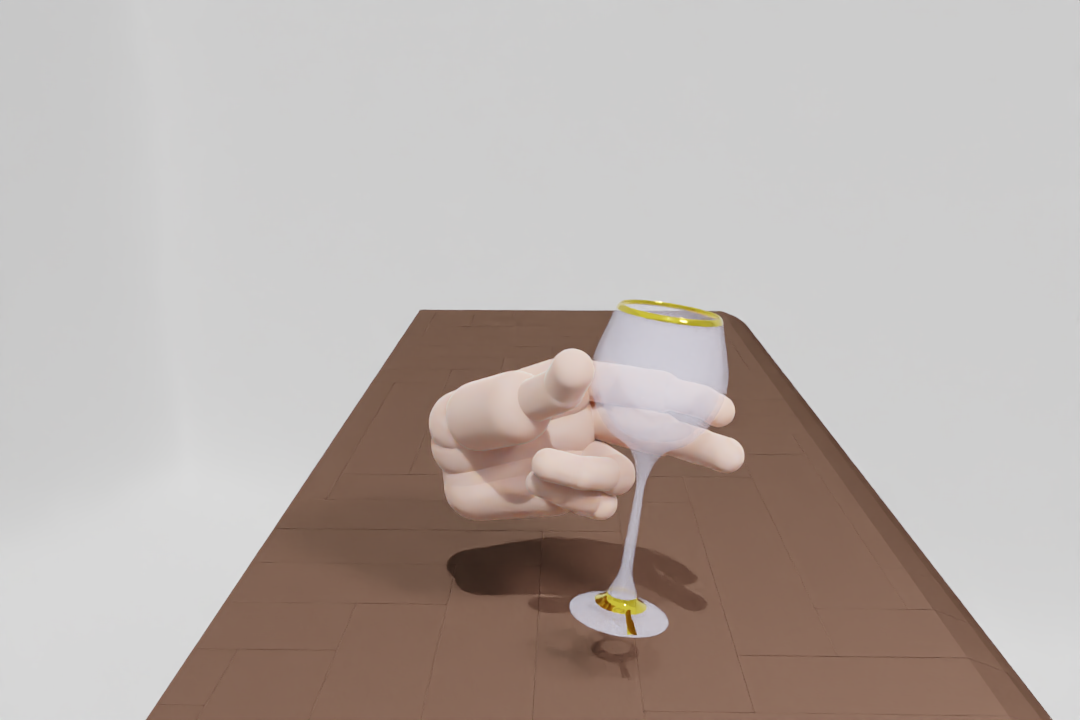}
    \hfill
    \includegraphics[height=\high]{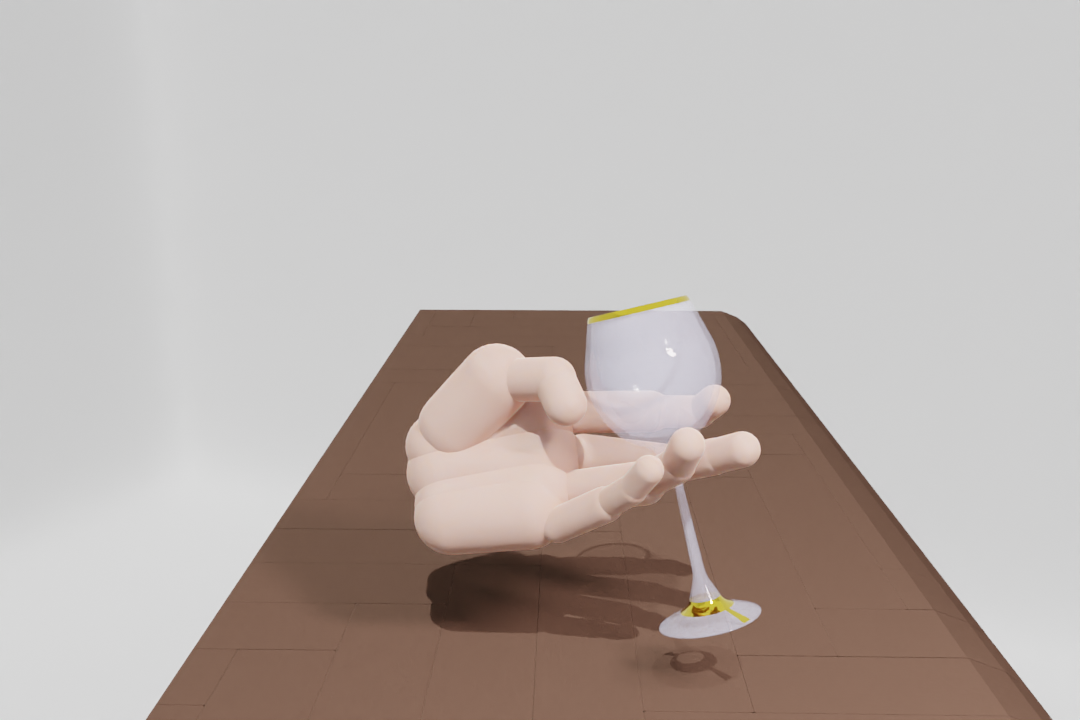}
    \hfill
    \includegraphics[height=\high]{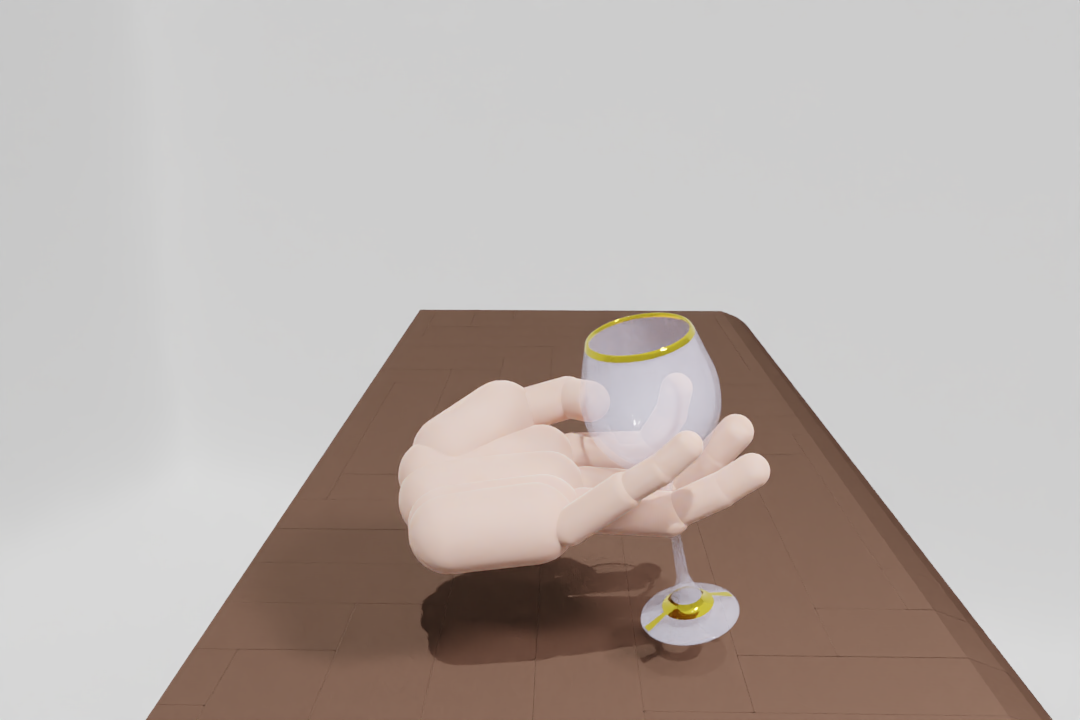}
    \hfill
    \label{fig:torus_large1_single}
\end{figure*}

\begin{figure*}[h!]
    \centering
    \includegraphics[height=\high]{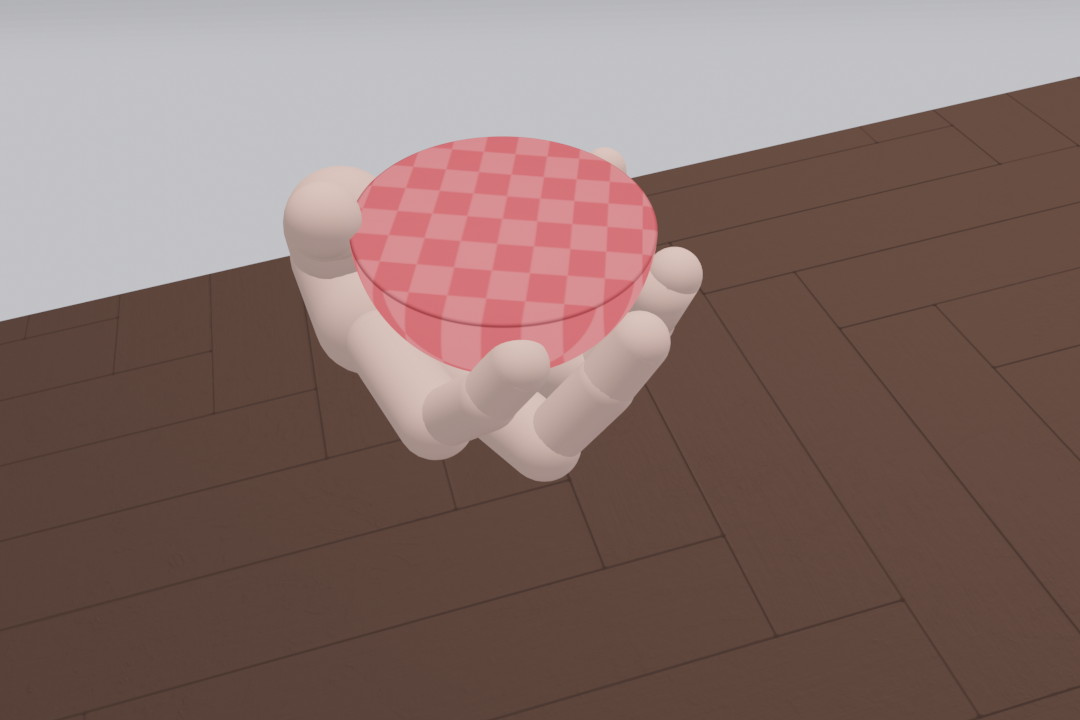}
    \hfill
    \includegraphics[height=\high]{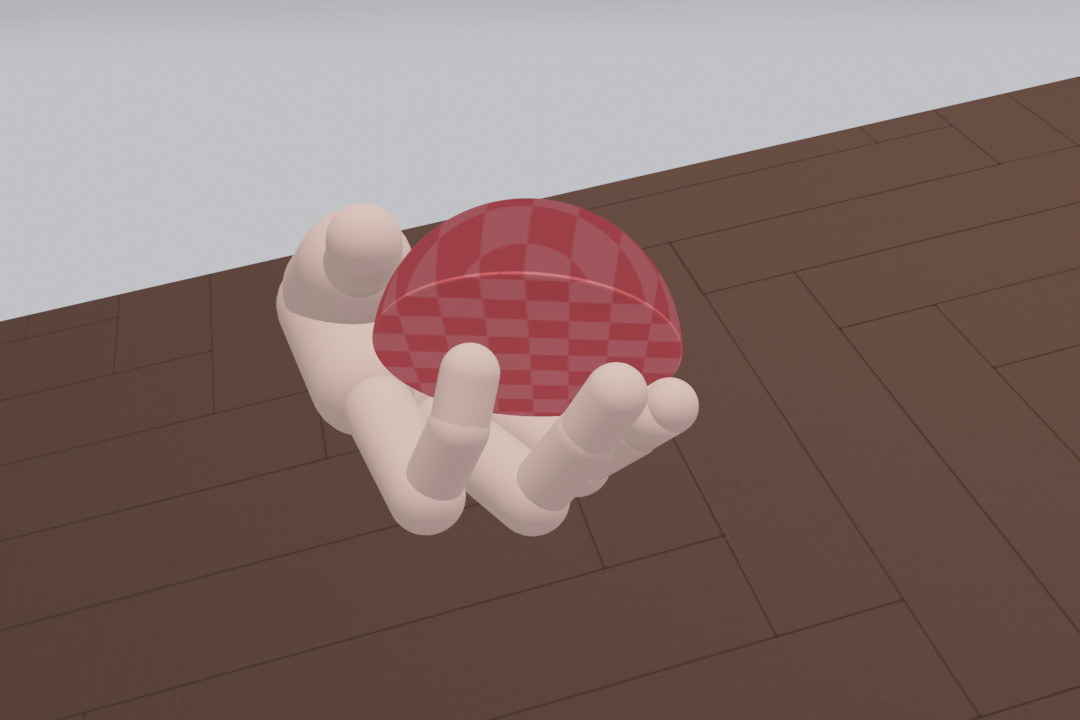}
    \hfill
    \includegraphics[height=\high]{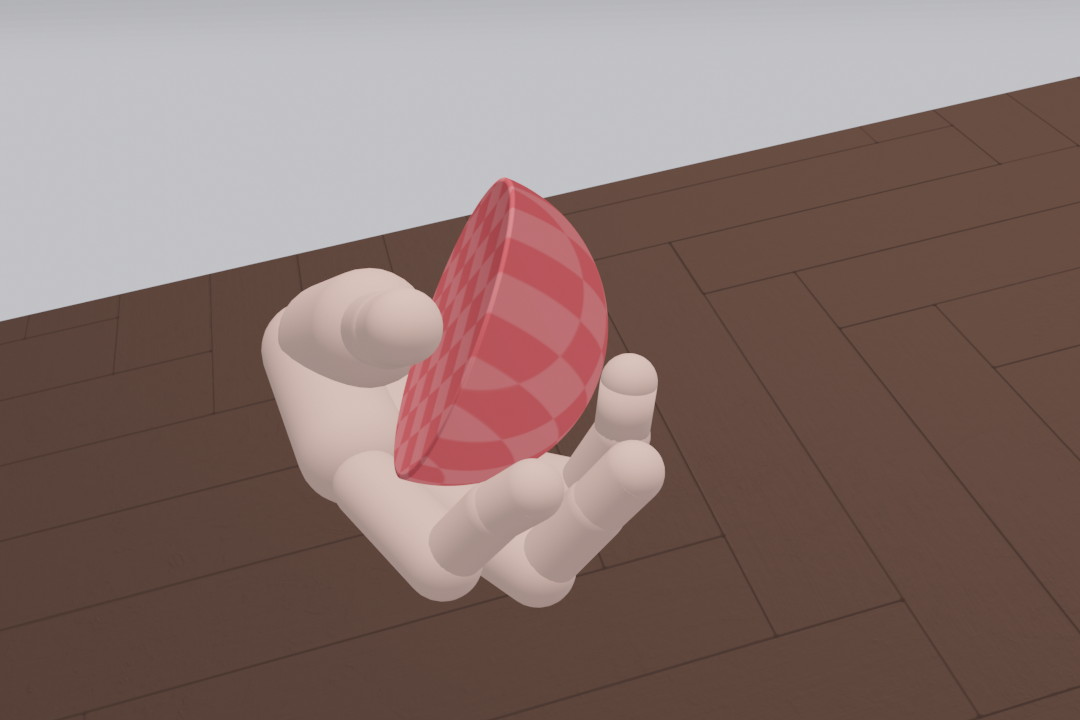}
    \hfill
    \includegraphics[height=\high]{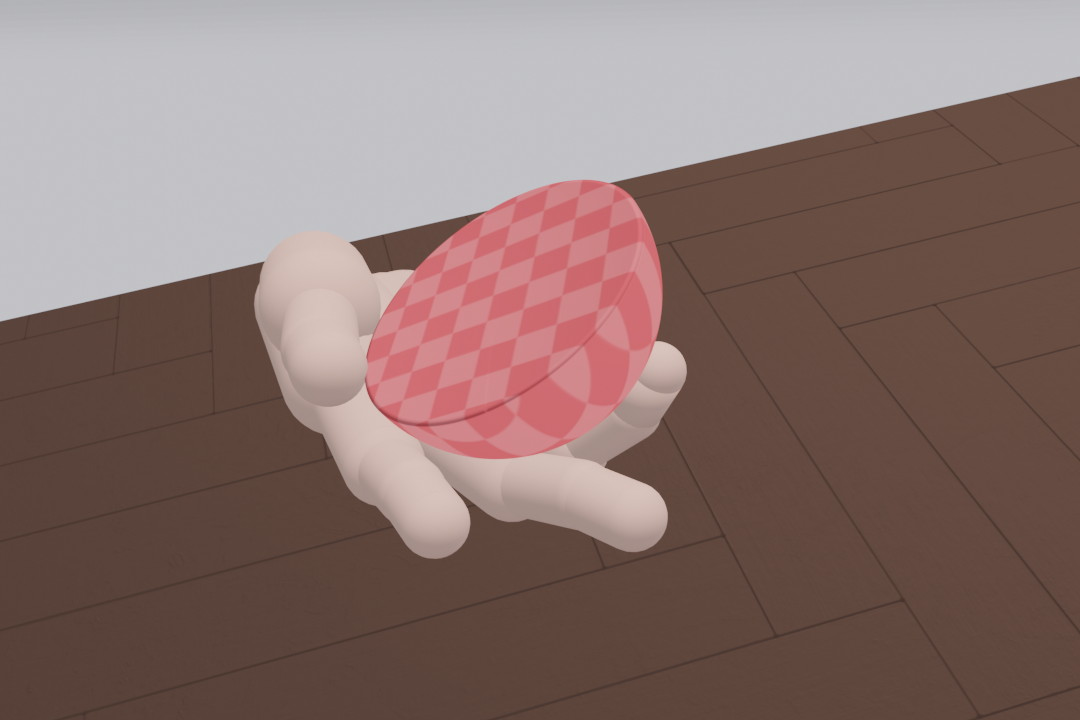}
    \hfill
    \includegraphics[height=\high]{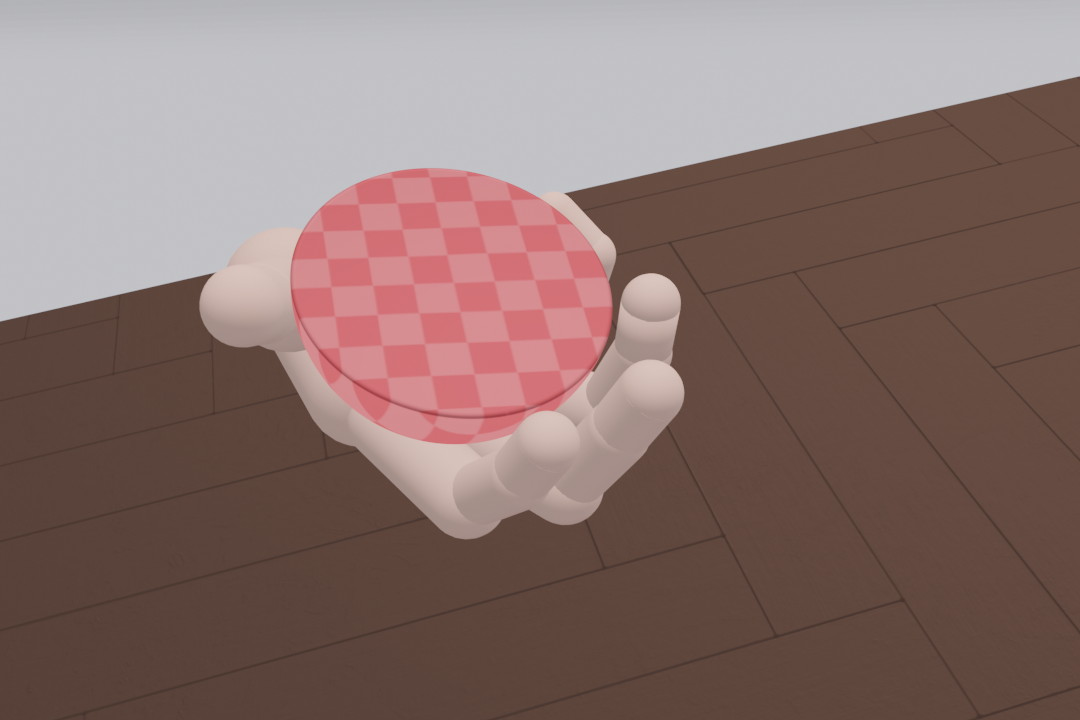}
    \hfill
    \caption{Still frames from manipulation sequences involving a wineglass (\textbf{top}) and a hemisphere (\textbf{bottom}).}
    \label{fig:hemisphere_large1_single}
    \vspace{-0.1in}
\end{figure*}
\begin{algorithm}[tb]
\caption{Greedy Shape Curriculum}
\label{alg:shape_morph_training}
\begin{algorithmic}[1]
    \STATE Initialize score list $S$
    \STATE Initialize policy list $\Pi$
    \STATE Initialize current shape $s=0$ and current policy $\pi = \Pi[s]$
    \FOR{$i = 0, 1, 2 ,\dots$}
    \IF{$i \mod k ==0$}
    \FOR{Every shape $j$}
    \STATE score = rollout on $j$ using policy $\pi$
    \IF{score > $S[j]$}
    \STATE $S[j]$ = score
    \STATE $\Pi[j] = \pi$
    \ENDIF
    \ENDFOR
    \STATE $s$ = Get best unsuccessful shape
    \STATE $\pi = \Pi[s]$
    \ENDIF
    \STATE PPO using $s$ and $\pi$
    \ENDFOR
\end{algorithmic}
\end{algorithm}

A key component of our greedy shape curriculum is a \emph{goodness score} that describes how likely a policy will succeed on a given shape. This metric will be used to update the best policy for a shape, and for selecting the next (shape, policy) pair for training. An obvious choice would be the average episodic reward. However, the consideration here is slightly different. A high episodic reward imposes a more strict constraint to the quality of object pose matching, making it hard to achieve when the target shape is too different from the source shape. A low episodic reward, on the other hand, cannot guarantee the completion of a rollout. On the other hand, the rollout length alone is too simple and fails to reflect the quality of the motion. We want a criteria with high tolerance to object deviation but with low tolerance to failure of completion. To this end, we design our criteria as a combination of the rollout duration and tracking accuracy, and this works robustly in practice. As described in \revised{Equation} \ref{equ:eval_score}, we use the product between the normalized episode length and the sum of hand joint reward of the rollout as the goodness score of a policy for a given shape. This encourages the resulting policy to use a similar manipulation strategy to the input. We consider a score higher than $d=0.55$ as \emph{successful}, and we only pick from the unsuccessful shape morphs for policy training to make progress.
\begin{equation}
    \label{equ:eval_score}
    f = \frac{L}{T} \cdot \frac{\sum_{0}^{L}{r_{joint}}}{T}
\end{equation}

Our greedy schedule is effectively an exploitation strategy, and we still need to balance it with some exploration to avoid local minima. If a particular shape is repeatedly picked for training and starving other shapes, we instead randomly select another shape and its paired policy in the next iteration. This is especially helpful when a policy gets ``stuck'' on a challenging frame towards the end of a sequence while other shapes have not had much training yet. Training progress on other shapes can then help improve such challenging cases later. Similarly, if we are successful on all shapes before the compute budget has been reached, we randomly pick a policy to continue training for further improvement. 

\section{Results}

\def \fw {0.18}
\begin{figure*}[h!]
    \centering
    \includegraphics[width=\fw\textwidth]{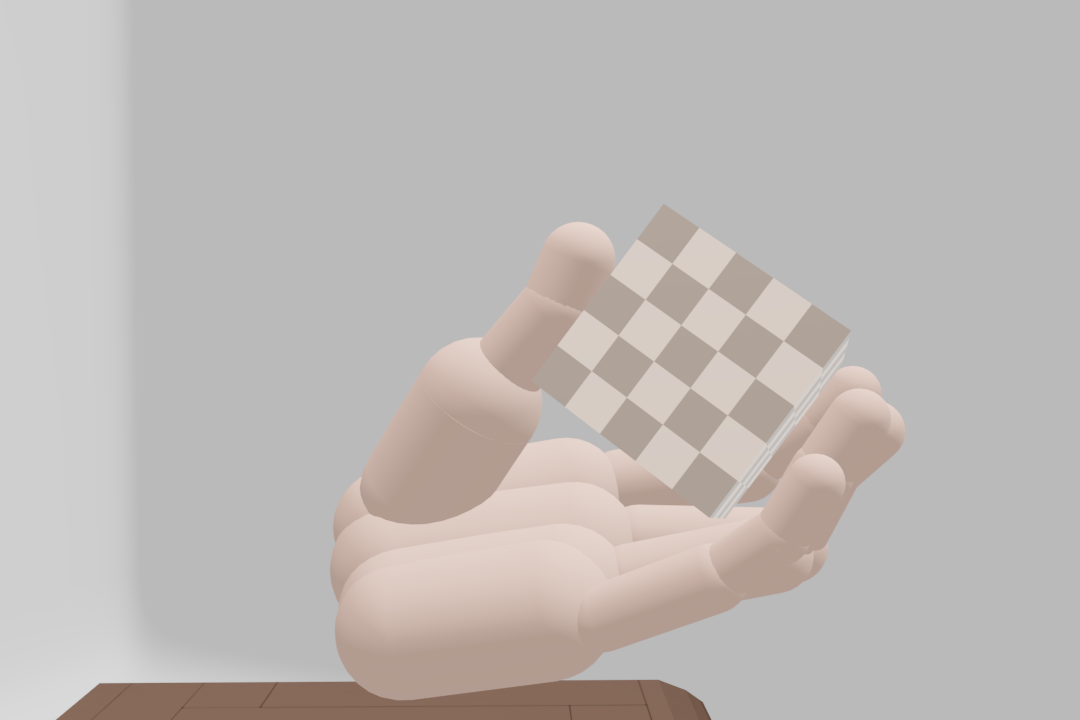}
    \includegraphics[width=\fw\textwidth]{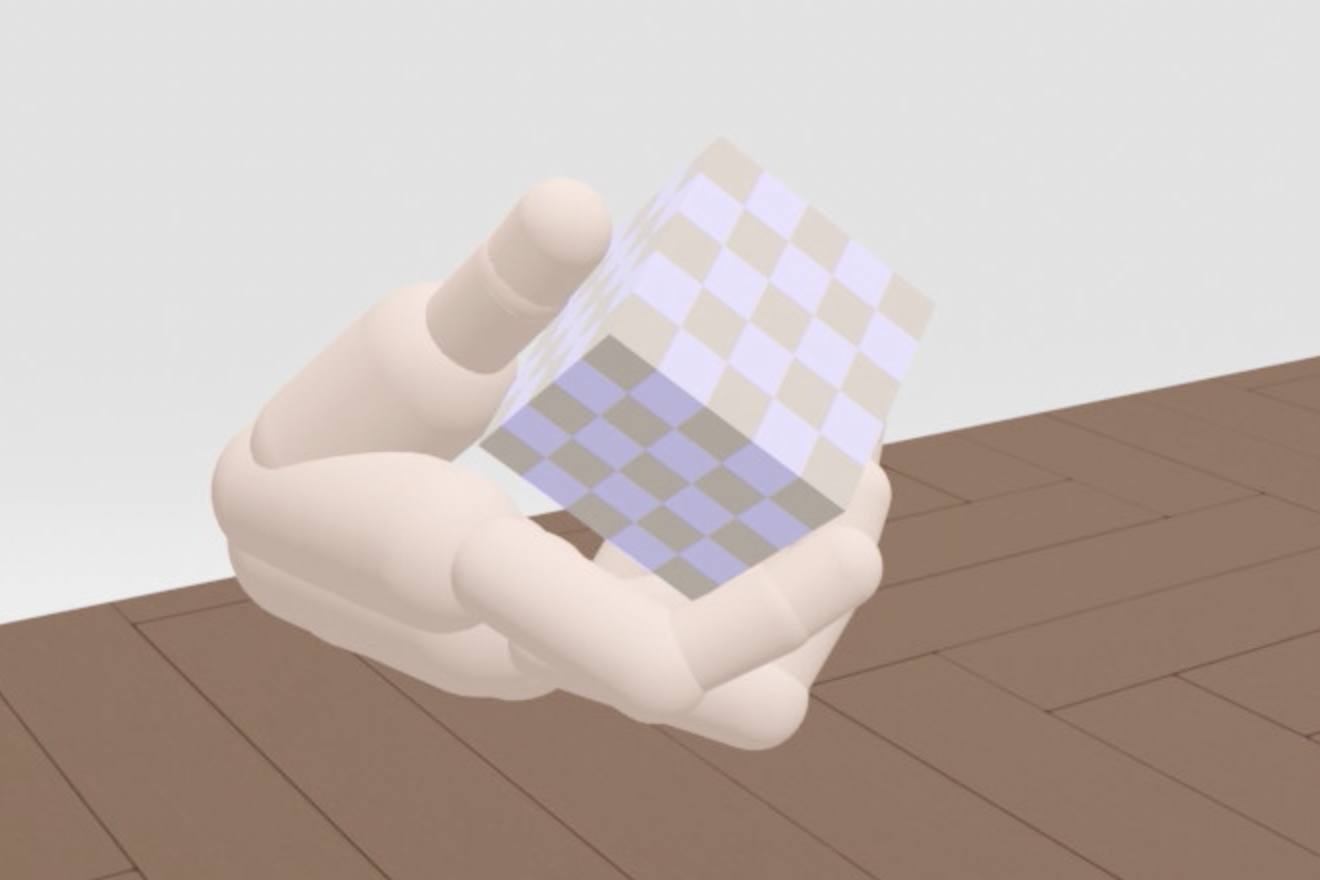}
    \includegraphics[width=\fw\textwidth]{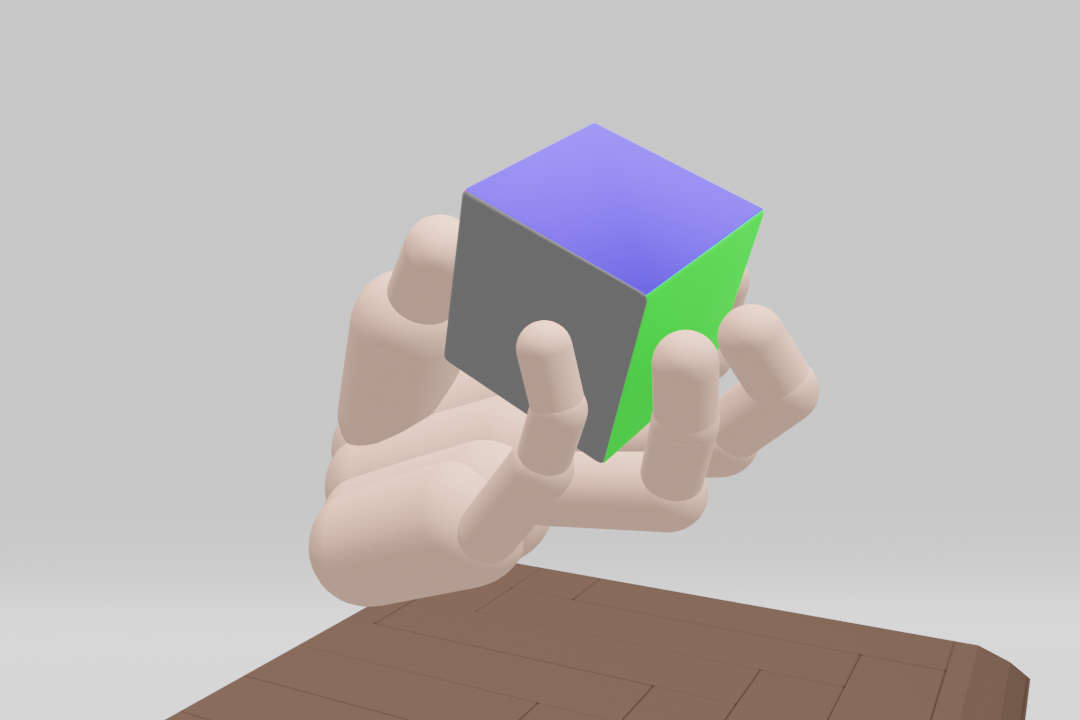}
    \includegraphics[width=\fw\textwidth]{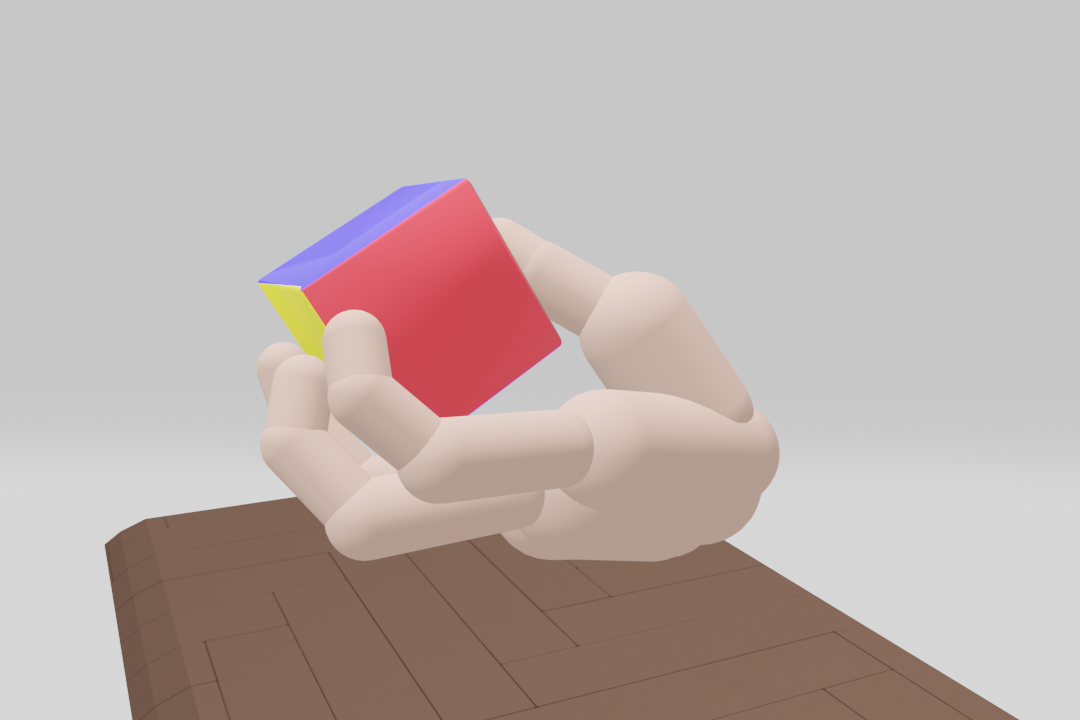}
    \includegraphics[width=\fw\textwidth]{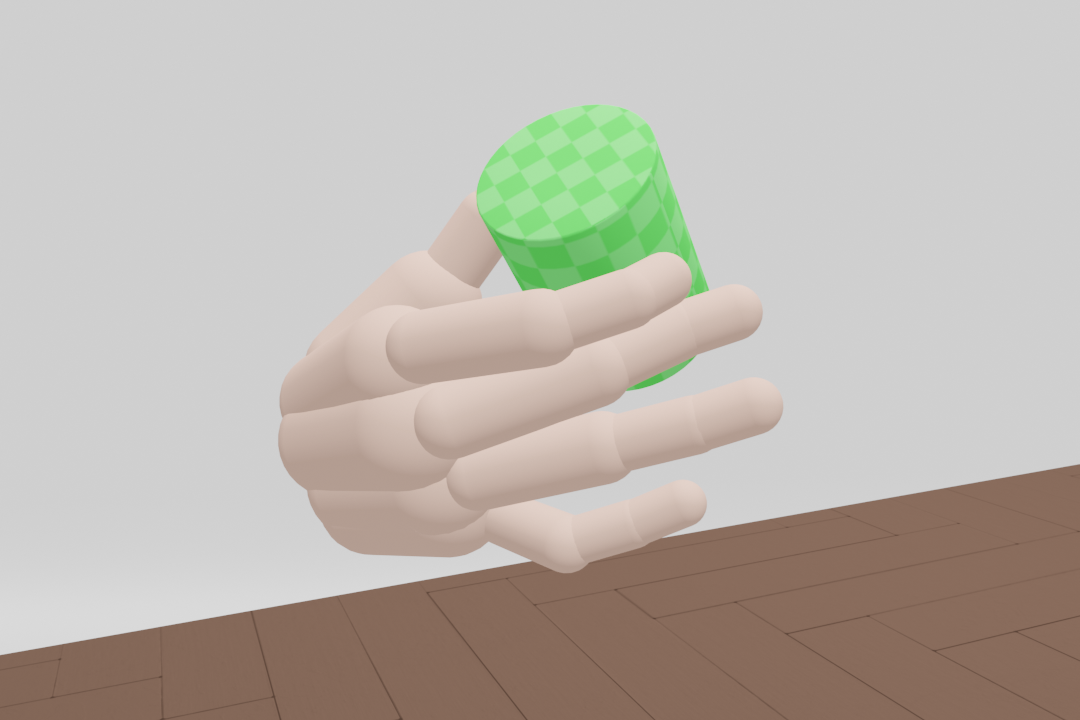}
    \begin{subfigure}{\fw\textwidth}
    \includegraphics[width=\textwidth]{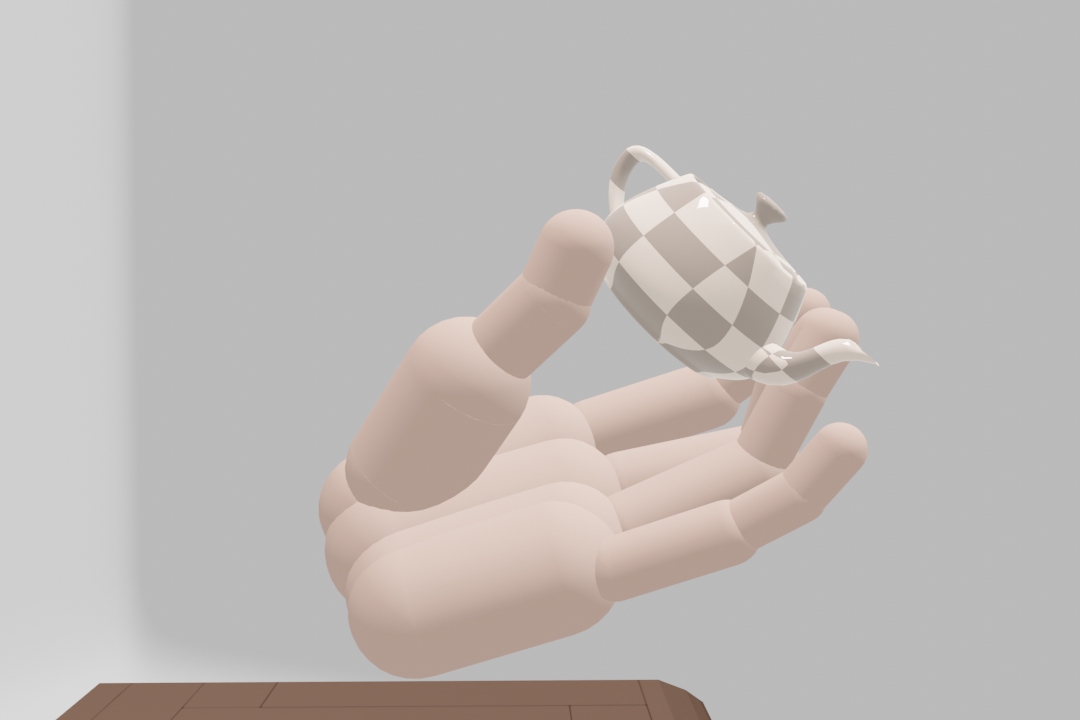}
    \caption{}
    \end{subfigure}
    \begin{subfigure}{\fw\textwidth}
    \includegraphics[width=\textwidth]{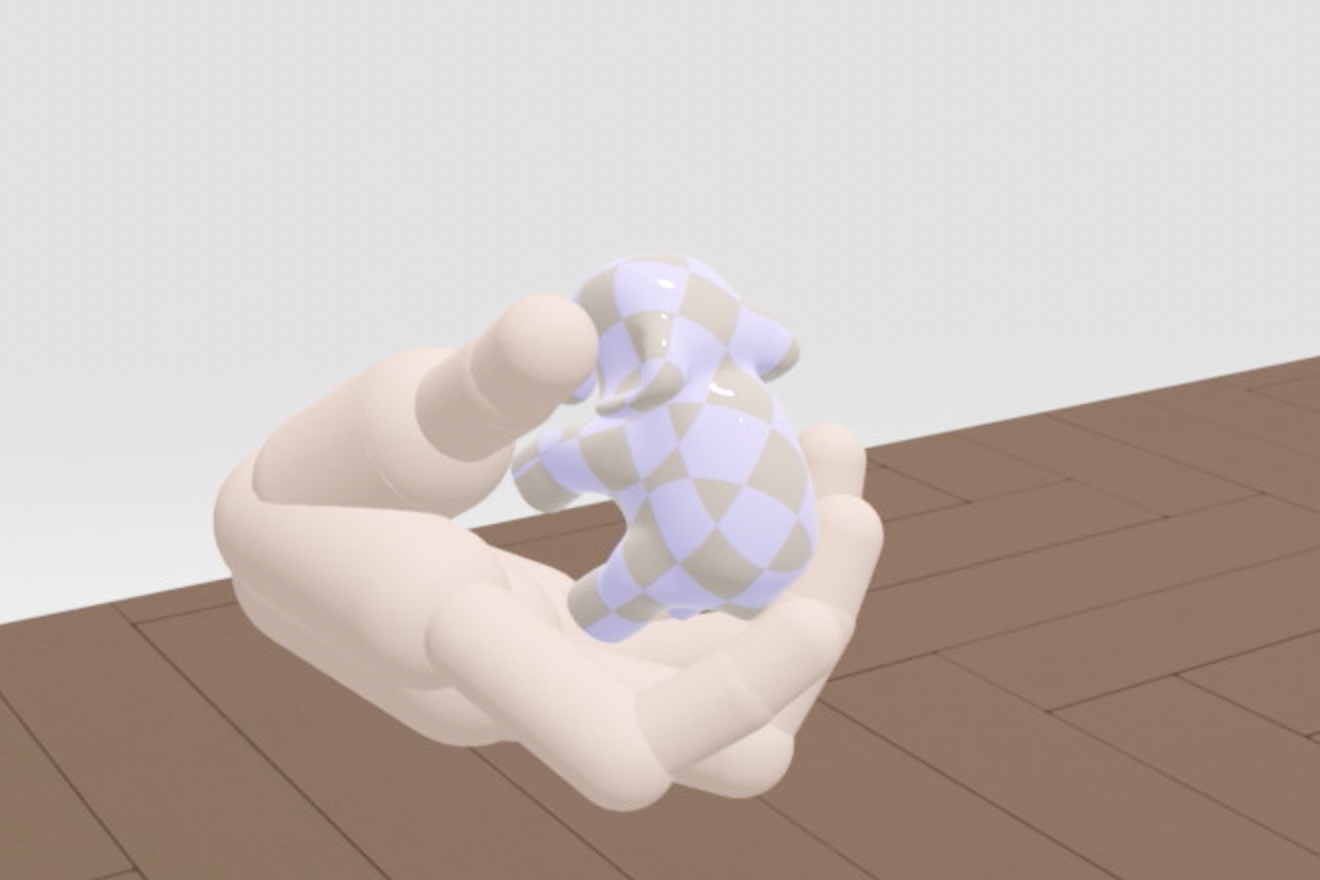}
    \caption{}
    \end{subfigure}
    \begin{subfigure}{\fw\textwidth}
    \includegraphics[width=\textwidth]{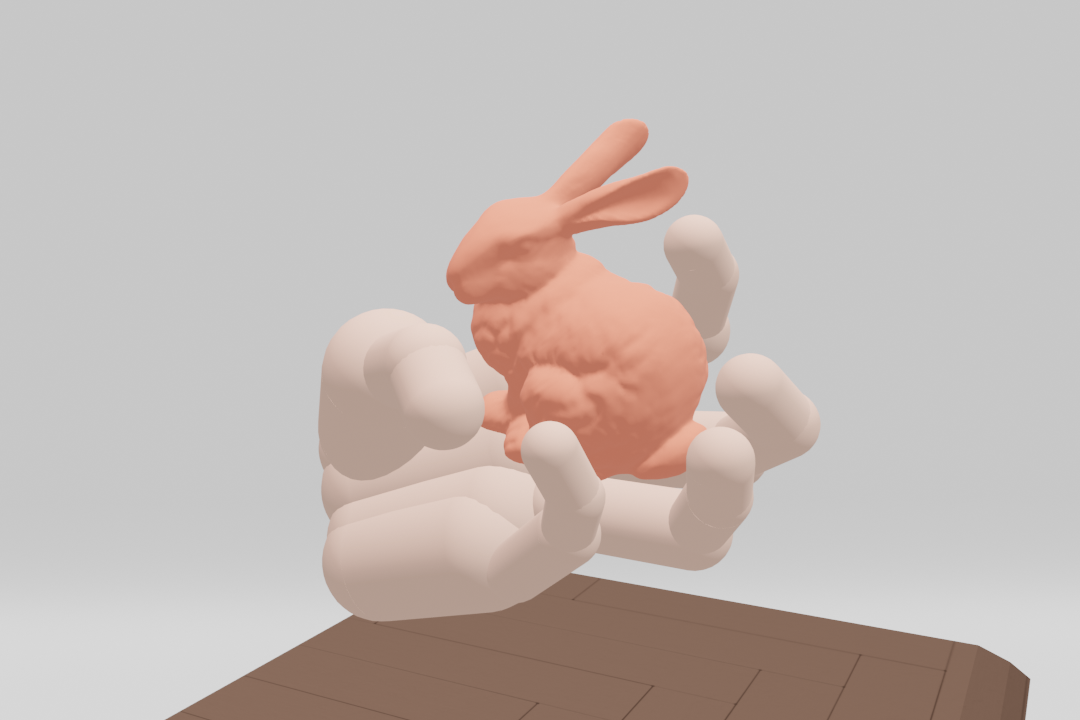}
    \caption{}
    \end{subfigure}
    \begin{subfigure}{\fw\textwidth}
    \includegraphics[width=\textwidth]{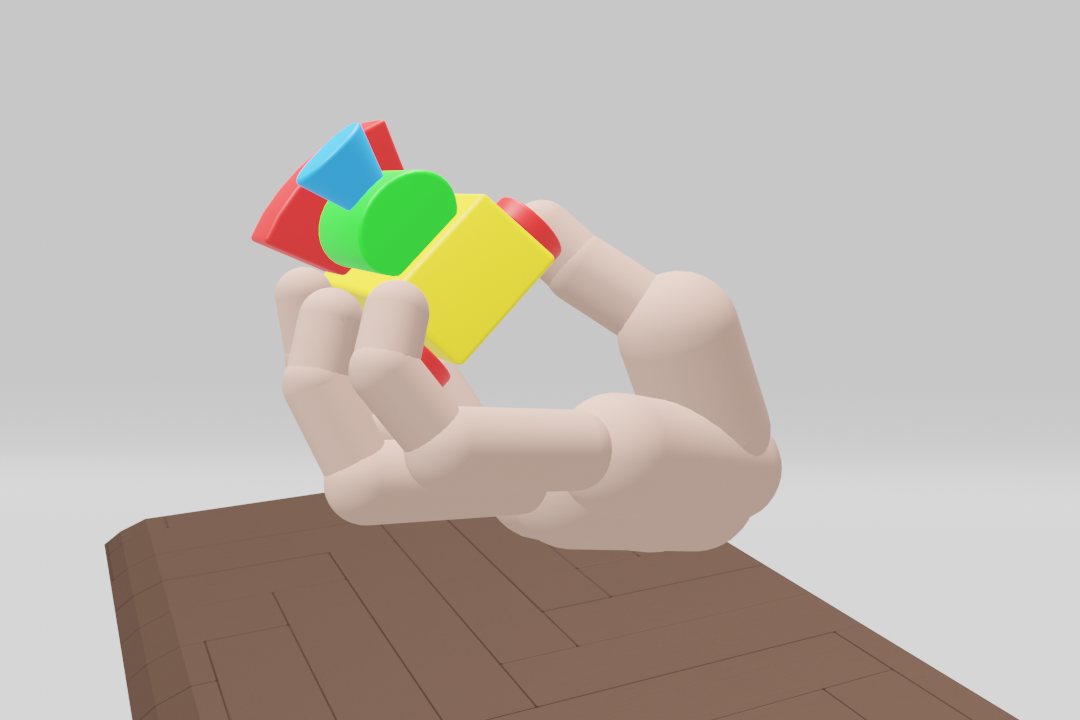}
    \caption{}
    \end{subfigure}
    \begin{subfigure}{\fw\textwidth}
    \includegraphics[width=\textwidth]{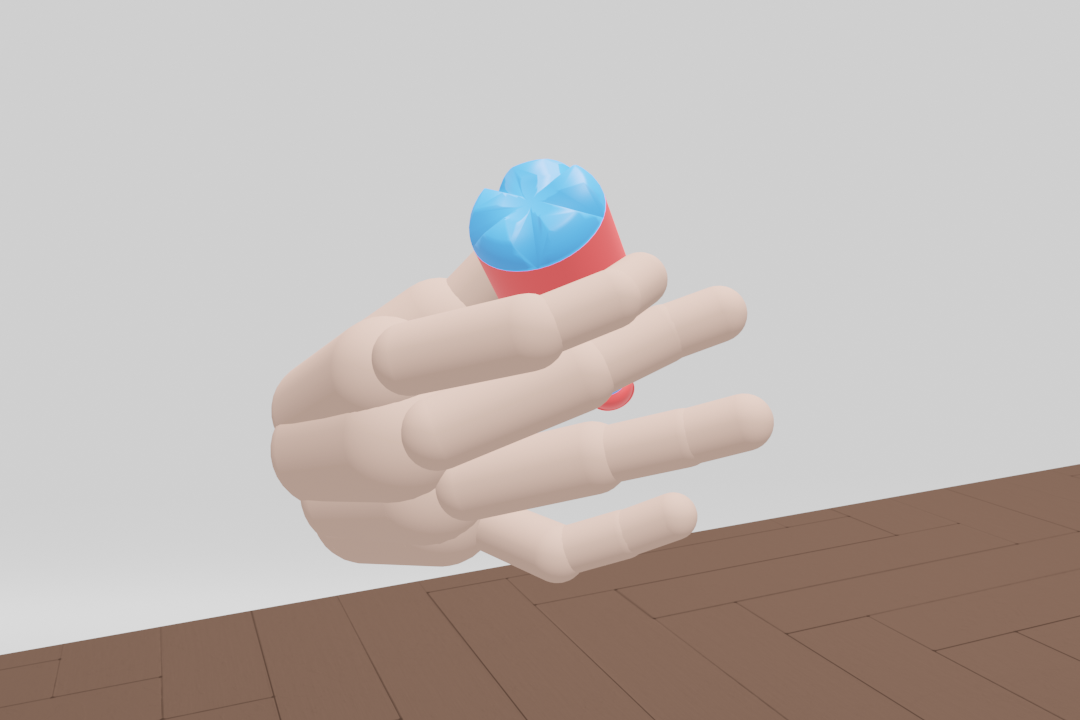}
    \caption{}
    \end{subfigure}
    \caption{Examples of transferring the manipulation of an original shape (top) to a target shape (bottom). Left to right, the examples shown are: cube to teapot, cube to elephant, cube to bunny, cube to toy train, and cylinder to bottle.}
    \label{fig:greedy_shape_curriculum}
\end{figure*}

To validate our approach, we evaluate the performance of our policy on a variety of mocap sequences. These include single-handed manipulations of several distinct objects, as well as passing and rotating an object between two hands. We show that the resulting policies can endure a moderate amount of dynamic perturbations. We also demonstrate successful transfer of manipulations to novel everyday objects using our greedy shape curriculum. Please refer to the accompanying video for visual evaluation.

For all of our results, we use Proximal Policy Optimization (PPO) \cite{schulman2017proximal}, a common on-policy reinforcement learning algorithm. The policy architecture is a fully connected network with two hidden layers, each of which has $256$ units with $tanh$ activation functions. Each policy is trained with 32 million observation/action pairs over 800 iterations. This training setting is held consistently across all trainings for original object sequences and shape morphs.

Because our source shapes are convex, we can use a simple projection-based shape morphing technique in order to generate intermediate shapes. To create the morphs, we project all vertices from the target shape to the surface of the source shape. We then use the intermediate positions along the projection paths as the vertex positions for each morph. For each shape pair, we create four intermediate morphs with vertices linearly interpolated between the corresponding positions on the source and target shapes. To create Mujoco compatible models, we use the commonly used convex decomposition method V-HACD\cite{mamou2016volumetric} with a maximum voxel resolution of $40 \times 40 \times 40$ to generate a collection of convex parts for each of the intermediate morphs. All the morphing operations are retrieved from a pre-processing stage in Blender~\cite{blender}. Figure \ref{fig:collision_shape_morphs} shows an example of the collision shapes of the source, target, and all intermediate morphs between a cube model and a bunny model. In later sections, we refer to each shape by its interpolation distance from the original shape along with the shape name. For any intermediate shape, the shape name is referred to as "Morph". For example, the sequence of morphs from the cube to the bunny are denoted as Cube(0.0), Morph(0.2), Morph(0.4),Morph(0.6),Morph(0.8), and Bunny(1.0).

We conduct our experiments on AWS C4 compute nodes with $36$ virtual cores on each machine. Each learner process executes $800$ iterations of PPO, collecting $32$ million rollout samples in total with $8$ parallel worker processes at a rate of $350$ frames per second. This results in an end-to-end policy training duration of about $40$ hours, depending on the complexity of the object's geometry and the number of contacts.



\subsection{Reproducing single dynamic sequences}
We first demonstrate the success of our approach across a diverse set of sequences involving a variety of objects and manipulation skills. First, we train a set of policies to track single-handed manipulation sequences, involving different objects over a $4$-second horizon. These results indicate that our approach is successful for sequences involving both primitive convex objects such as cubes, cylinders, and hemispheres, as well as more complex, concave geometries such as a torus and a wineglass. Figure~\ref{fig:hemisphere_large1_single} (\textbf{top}) shows still frames from the manipulation of a wineglass, an example of a non-convex object, and (\textbf{bottom}) shows still frames from rotating a hemisphere in the hand. Additionally, we show that our approach can be trivially extended to control two-handed manipulations that require coordination, such as passing and cooperative rotation shown in Figure ~\ref{fig:teaser}.

\begin{table}[b!]
\vspace{3mm}
\caption{Success rate on example sequences with stochastic policy execution (500 samples). The maximum cumulative reward is 500.}
\vspace{-3mm}
\begin{center}
\begin{tabular}{|c|c|c|c|}
\hline
Experiment & Sequence & Success \% & Cumulative Reward \\
\hline
\multirow{9}{*}{\makecell{Single Hand }}& Cube 1 & 94 & 463.16\\
\hhline{~---}
& Cube 2 & 81.6 & 415.18\\
\hhline{~---}
& Cube 3 & 98.6 & 480.45\\
\hhline{~---}
& Cylinder 1 & 75.6 &  403.88\\
\hhline{~---}
& Hemisphere 1 & 88.6 &  431.58\\
\hhline{~---}
& Torus 1 & 85 &  432.55\\
\hhline{~---}
& Wineglass 1 & 94.2 &  458.99\\
\hline
\multirow{2}{*}{\makecell{Two Hand}} &Cube Passing & 82.6 & 417.66 \\
\hhline{~---}
&Cube Rotation & 99.6 & 482.63 \\
\hline

\end{tabular}
\end{center}
\label{tbl:success_rate_spcecialized}
\end{table}

Even though each policy is trained to follow one object manipulation sequence, it is able to endure moderate dynamic perturbations. To demonstrate this, at random frames through the simulation, we apply forces of $8N$ on all fingertips for $0.25$ second. Despite these perturbations, the policies adapt to these forces and still successfully track the motion. Similarly, a policy can easily track a motion sequence when modifying the masses and friction coefficients of the manipulated objects. These results are best seen in the supplementary video.

\begin{figure}[t]
    \centering
    \includegraphics[width=\linewidth]{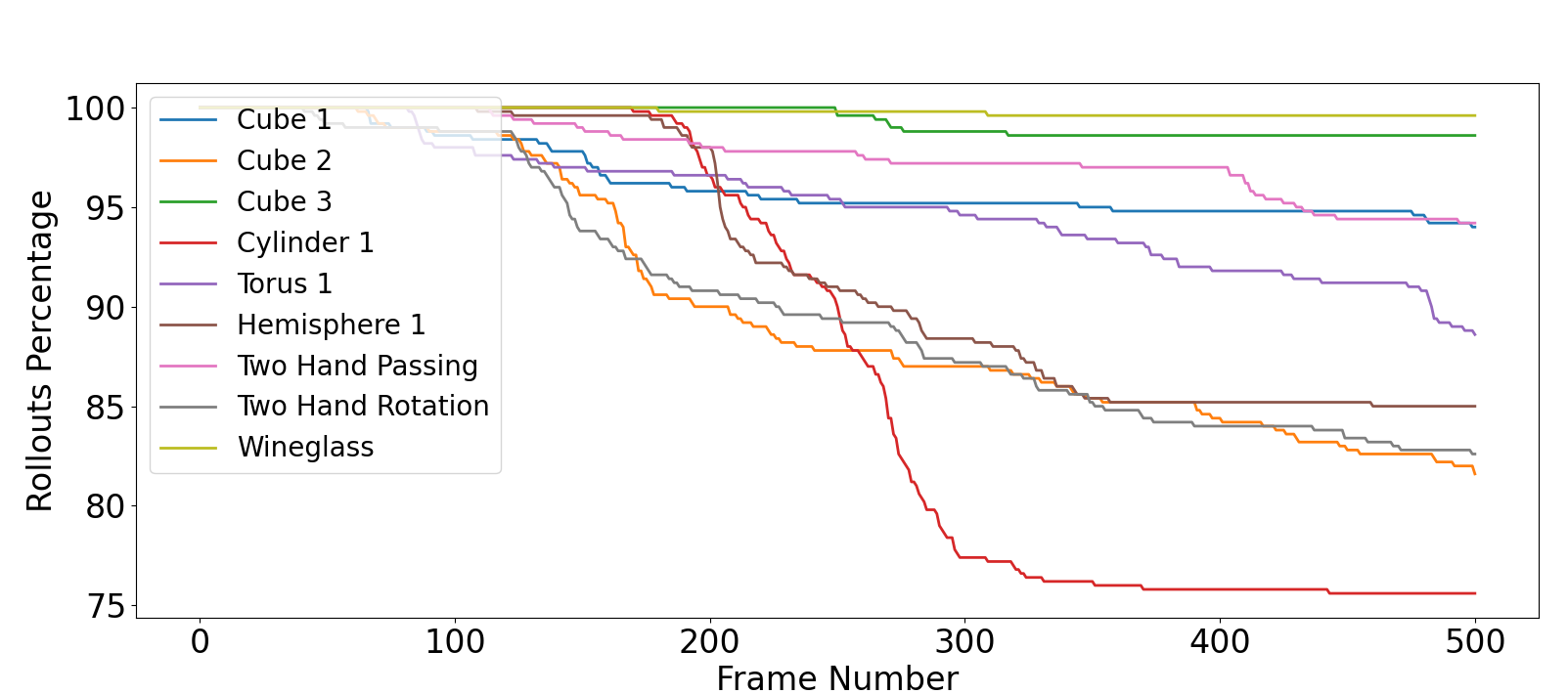}

    \caption{Completion percentage of all sequences. This plot illustrates the relationships between the difficulty of a given frame and the associated success rate. The least performing sequence (Cylinder1) has more than 75\% success rate to complete the entire rollout.}
    \label{fig:no_variation_threshold}
    \vspace{-0.18in}
\end{figure}

We summarize a quantitative analysis of success rates and mean reward of our final policies across the full set of motions in Table~\ref{tbl:success_rate_spcecialized}. To compute each entry in the table, a sample population of $500$ episode rollouts are initialized from the first frame of each motion capture sequence, and stochastic policy actions are applied until the final frame is reached successfully or the object tracking error crosses a threshold value. For a more detailed analysis and comparison, we cache the episode length of each trial, and plot the percentage of the rollouts that meet or exceed a range of rollout length thresholds.
In our experience, a success rate was too simple to capture the performance of the policy for our problem because it does not capture various difficulties of different frames. To this end, we plot ``completion percentage'' that shows the percentage of successful policies (Y-axis) to reach the given frame (X-axis). We summarized completion percentages in Figure \ref{fig:no_variation_threshold}. The figure shows how difficult each sequence is, and how often a final trained policy can successfully track the object to the end of the sequence. 
For instance, the cylinder 1 motion shows a significant drop of the completion percentage around frame $300$, where thumbs are having trouble rotating a standing cylinder back into the palm, and thus where the cylinder may fall out of the hand.

\begin{figure*}
\centering
\includegraphics[width=0.46\textwidth]{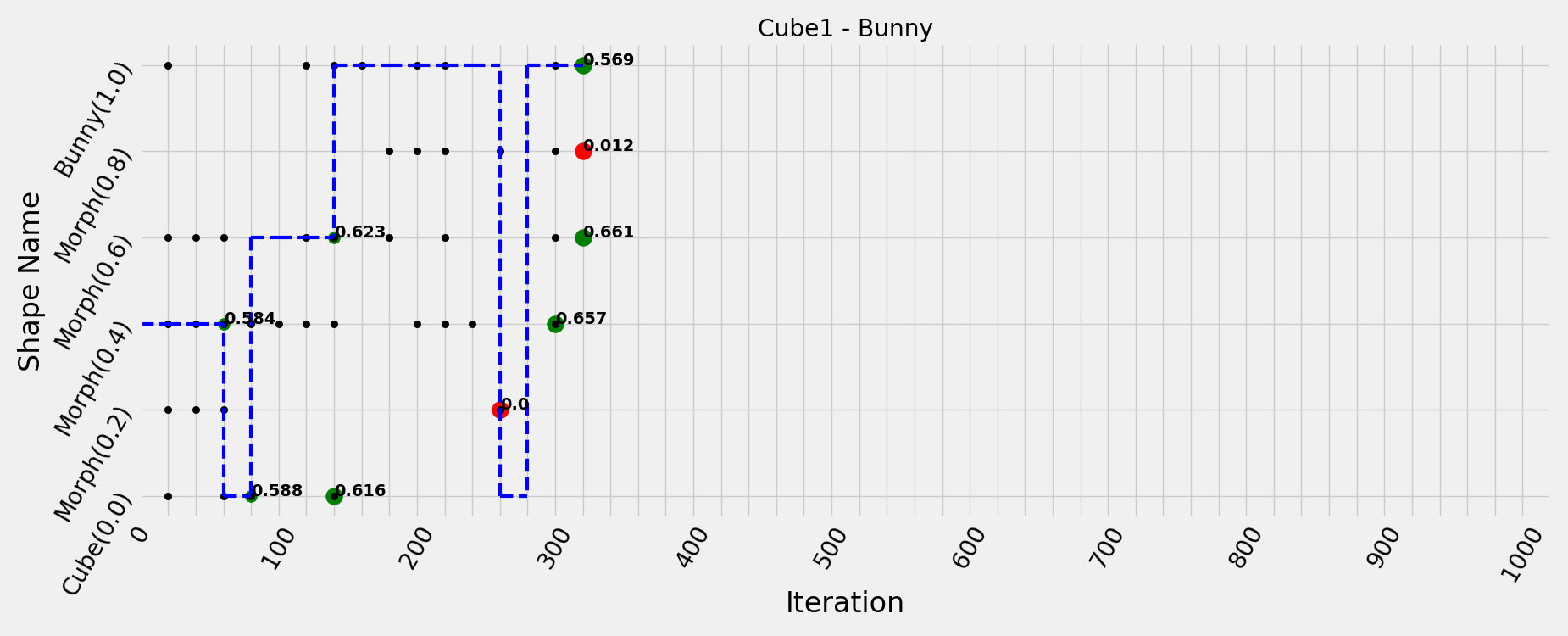}
\includegraphics[width=0.46\textwidth]{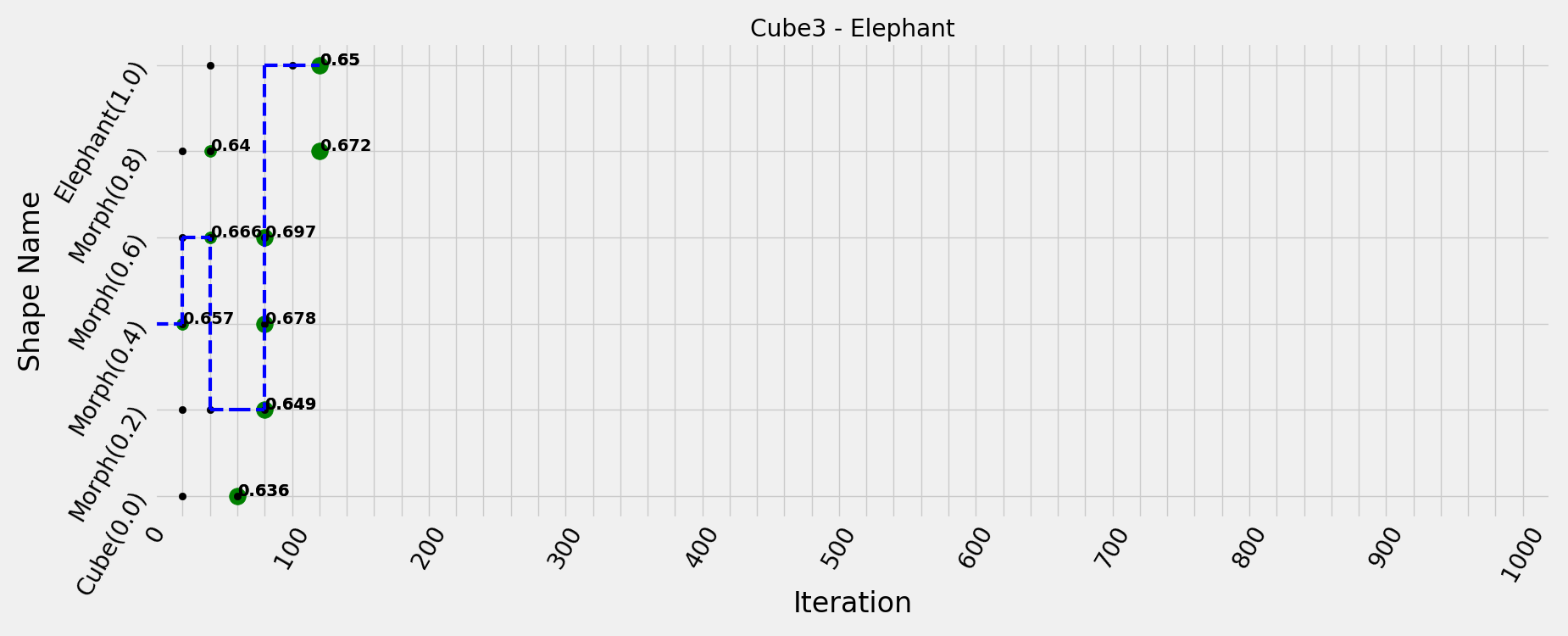}
\begin{subfigure}{0.46\textwidth}
\includegraphics[width=\textwidth]{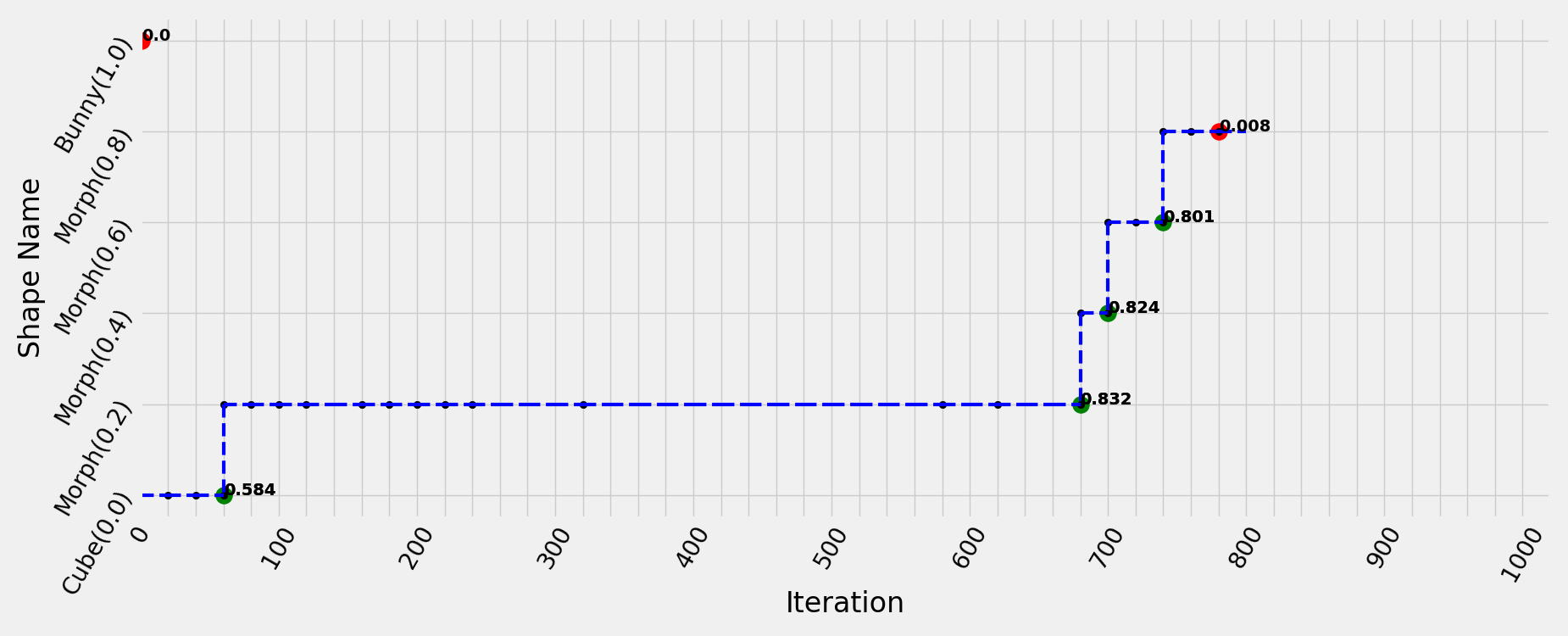}
\caption{}
\end{subfigure}
\begin{subfigure}{0.46\textwidth}
\includegraphics[width=\textwidth]{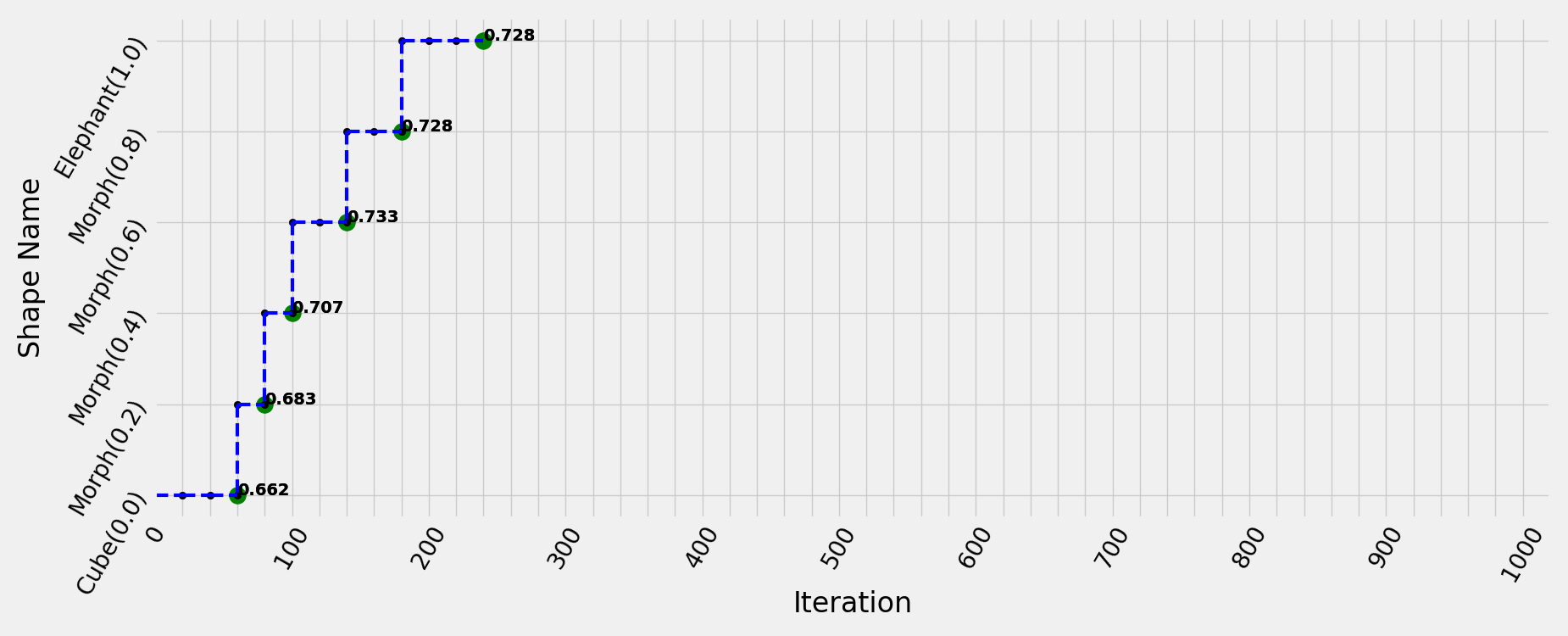}
\caption{}
\end{subfigure}

\caption{
Comparison between our greedy shape curriculum (top) and naive curriculum (bottom) method. Part (a) shows policy training for cube-to-bunny, and part (b) shows cube-to-elephant. Note that our method often trains more quickly than the naive approach. The graph at the bottom of (a) shows failure of the naive approach to even find a solution. The blue dashed lines show which shape is being trained at each iteration. Black dots indicate which shapes can receive higher goodness score when using the currently trained policy. Green dots indicate the first successful iteration with the corresponding goodness score, and red dots means the shape was not able to be solved when training completes.
}

\vspace{-0.18in}
\label{fig:dependency_plot_comparison}
\end{figure*}
\subsection{Transferring to Complex Objects using a Greedy Shape Curriculum}\label{ssec:policy_transferring}

We demonstrate that using our greedy shape curriculum, we can easily train policies that can manipulate both the source object and the target object by tracking a provided motion clip. For example, Figure \ref{fig:greedy_shape_curriculum} shows the simulation result after adapting the manipulation of a simple shape (e.g. cube or cylinder) to a more complex object (teapot, elephant, bunny, toy train, bottle).



We compare our method with a naive curriculum learning method where a policy will only proceed to the next shape morph when the previous one receives a goodness score above the threshold. Two examples are shown in figure~\ref{fig:dependency_plot_comparison}. In the figures, blue dashed lines indicate the policy training progression, and the black dots indicate which shapes are receiving a better score when evaluating the current policy. We highlight the first success of each shape in green dots labeled with the respective score. Greedy shape curriculum results are shown on the top row, and the naive curriculum results are on the bottom row.

Figure~\ref{fig:dependency_plot_comparison} (a) shows an example in which the intermediate morphs do not introduce a smooth linear curriculum.  The naive curriculum (bottom row) spends the majority of its compute budget on Morph(0.2) and eventually stops at Morph(0.8) without reaching the target shape. In contrast, the greedy shape curriculum (top row) skips training on these two difficult morphs, yet still finds a successful policy for the target shape by utilizing the other intermediate morphs. Training starts on Morph($0.4$) and finishes much sooner than the $800$ iteration budget at iteration $320$, by alternating among Cube(0.0), Morph(0.6), and the target Bunny(1.0). The two difficult shapes, Morph(0.2) and Morph(0.8), do not get trained on and are not successful at the end.



Our greedy shape curriculum can also be more efficient than the naive curriculum in easy cases when both methods succeed. In Figure~\ref{fig:dependency_plot_comparison} (b), the greedy shape curriculum picks the target shape Elephant(1.0) for training at iteration $80$ by inheriting the policy trained on Morph($0.6$) at iteration $20$. After further training, it is able to successfully manipulate the elephant at iteration $120$. On the other hand, the naive curriculum has to receive high goodness score on all five previous shapes before succeeding on the target shape Elephant(1.0) at iteration $240$. 

To highlight the benefits of our method, we compare greedy shape curriculum with four different baselines: 1) Training a single policy on the target shape; 2) Training a single policy over both the source and the target shapes; 3) Training a single policy over the collection of morphs from the source and the target shapes; 4) Training a collection of policies using a naive curriculum over all morphs. For each of the baselines, we take the mean policies trained from four random seeds, apply them on the target shape, and record whether or not they can successfully complete the rollout without hitting the early termination criterion. All trials are trained using 32 million samples over 800 iterations. As shown in Table ~\ref{tbl:baselines_comp}, Greedy Shape Curriculum has the highest likelihood of producing working policies for all the target shapes under a fixed sample budget. In comparison, training on the target shape directly has a lower chance of success. Using a naive curriculum may be helpful in some cases but could be harmful in others, because only a subset of the sample budget is allocated to improve the target shape. On the other hand, training one generalized policy on multiple shapes is making the problem more challenging and is therefore more difficult to succeed, especially when the compute budget is limited.


\begin{table*}[t!]
\vspace{3mm}
\caption{Comparing the success rates of our method against four different baseline methods. Each method is trained with four random seeds.}
\begin{center}
\begin{tabular}{|c|c|c|c|c|c|}
\hline
Original Sequence - Target Shape  & Direct Target & Source + Target & All morphs & Naive Curriculum & Greedy Shape Curriculum \\
\hline
Cube1 - Teapot & 50\% & 0\% & 0\% & 50\% & \textbf{75\%} \\
\hline
Cube1 - Bunny & \textbf{100\%} & 0\% & 0\% & 50\% & \textbf{100\%} \\
\hline
Cube1 - Train & \textbf{100\%} & 50\% & 0\% & 75\% & \textbf{100\%} \\
\hline
Cube2 - Bunny & 50\% & 0\% & 0\% & 0\% & \textbf{100\%} \\
\hline
Cube2 - Elephant & \textbf{25\%} & 0\% & 0\% & 0\% & \textbf{25\%} \\
\hline
Cube3 - Elephant & 50\% & 0\% & 0\% & \textbf{100\%} & \textbf{100\%} \\
\hline
\end{tabular}
\end{center}
\vspace{-0.15in}
\label{tbl:baselines_comp}
\end{table*}

\subsection{Refining successful policies}
\begin{figure}[h!]
    \centering
    \includegraphics[height=\high]{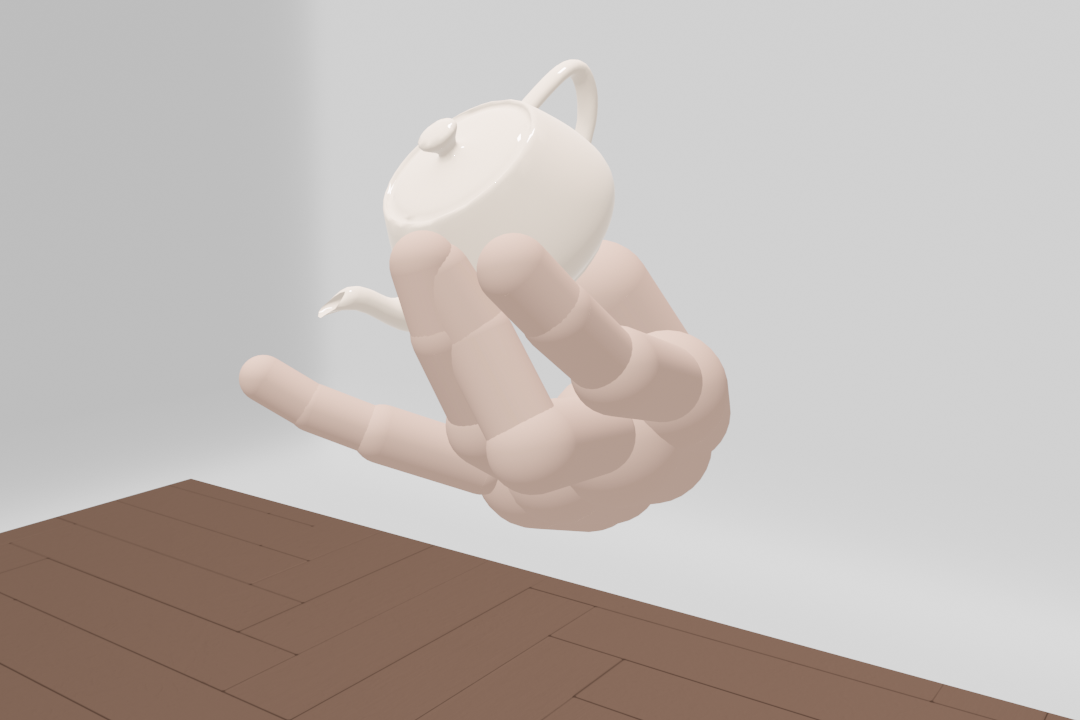}
    \hfill
    \includegraphics[height=\high]{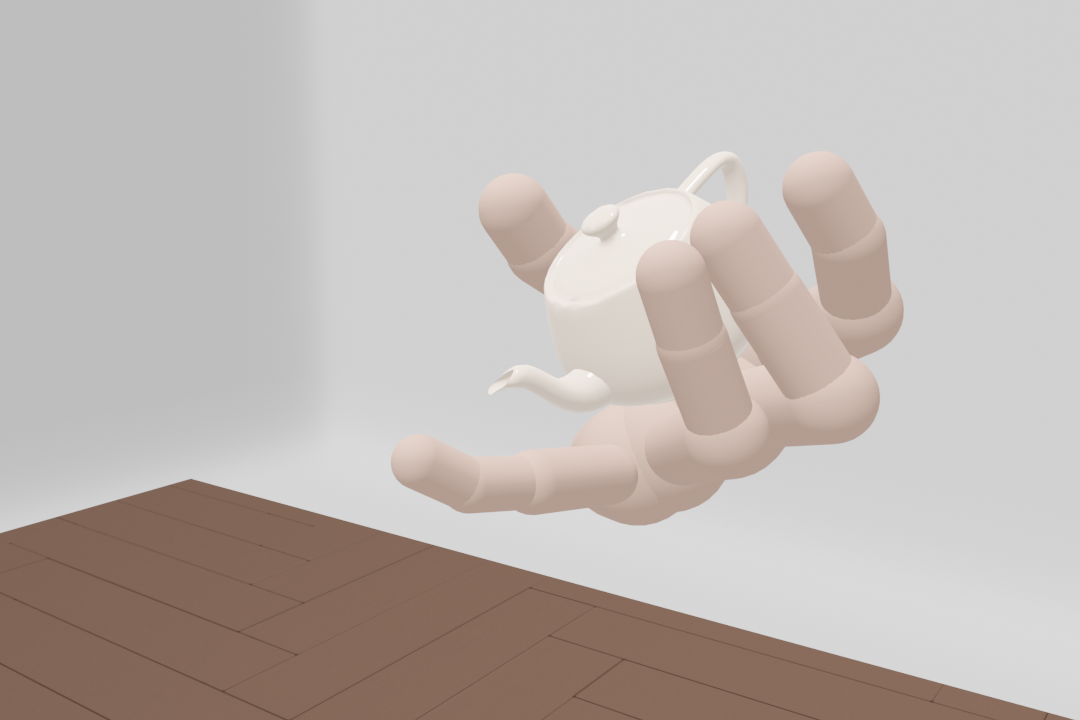}
    \caption{Interpenetration between fingers (\textbf{left}) is resolved after refining the policy with self-collisions enabled (\textbf{right}).}
    \label{fig:comparing_refinement}
    \vspace{-0.1in}
\end{figure}

\revised{Due to the simplification on disabling the self collision within the hand, the generated animation sometimes have the hand interpenetrating between the fingers. To mitigate the issue, we take the trained policy on target morph from the result of Greedy shape curriculum, and refine the training with a hand model with self collision enabled. This simple refinement can often result in more physically realistic motion qualities for the manipulation. A comparison is shown in Figure \ref{fig:comparing_refinement}}.

\section{Limitations}


Although we have been successful in tracking reference manipulations, we still sometimes observe noticeable penetrations between the hand and the object. Such artifacts are due to the contact handling mechanism in the physics engine. They are more pronounced when the object has a narrow or thin feature, such as the stem of the wine glass. 
\revised{In addition, we found that having a self-collision enabled hand makes the physics simulation very fragile that rollouts can easily get early terminated during training. Successful policies on target morphs can hardly be trained under this setup, and we end up disabling the self-collisions within the hand, and lead to some finger interpenetration in our results. Even though further refining the resulting policies may mitigate the artifacts, it's not guaranteed to reach an interpenetration-free rollout.}



We are still limited in how much the target shape can deviate from the source shape, and we cannot guarantee to always find a successful policy. With extreme changes in shape or size, the original manipulation strategy as captured may no longer be suitable or feasible. Related to this, our method cannot discover and explore novel manipulation skills that deviate significantly from the input. Even when successful (eg. the object is not dropped), the resulting finger motions may at times appear unnatural. The simple imitation term in the reward function and in the goodness score trades off exploration and exploitation, and could become a limiting factor in inventing new ways to interact with an object. An Adversarial Motion Prior~\cite{peng2021amp} could be a promising mitigation.

Lastly, we found that some mocap sequences are particularly difficult to simulate because of noisy or erroneous frames. Although the motion imitation can learn a robust policy and ``correct'' these invalid frames via physics simulation, they tend to significantly slow down the learning process.


\section{Conclusion}


We have demonstrated training of policies for in-hand manipulation of objects based on motion capture data. Moreover, using a greedy shape curriculum, we can also adapt a given motion to a new shape. Because we use physics simulation, any disturbances or changes in physical properties are reflected in the object and hand motions. Our results closely mimic the fluidity and naturalness of real hands in the mocap examples.


One interesting direction for future work would be to reuse the existing motion clips for generating novel manipulation motions.
Building a manipulation motion graph~\cite{lee2019scalable} to smoothly join manipulation mocap clips would be one possible approach. Unlike locomotion data where activities are highly repetitive, it is much harder to find suitable transition points with similar object and contact states, making this a challenging direction to explore.

\revised{Currently, policies generated from our method are only specialized to the target shape that it is trained on. We believe it is a first step towards solving a universal policy that can generalize to arbitrary shapes and task goals, which would be fruitful future work}

\section{Acknowledgement}
We acknowledge Noah Maestre for his help on preparing the rendering setup of the animations. This work is partially supported by Meta Reality Lab.
\bibliographystyle{eg-alpha-doi} 
\bibliography{references}       


\end{document}